\documentclass[11pt,a4paper,logo,copyright,nonumbering]{unimobileagent}

\usepackage[utf8]{inputenc}
\usepackage[T1]{fontenc}

\usepackage{xcolor}

\definecolor{citeblue}{rgb}{0.21,0.49,0.74}
\definecolor{headerbg}{HTML}{F2F5F7}
\definecolor{linkblue}{HTML}{006FE6}
\definecolor{brandblue}{HTML}{2E5AA8}
\definecolor{abstractpurple}{HTML}{8E5AC8} 

\usepackage{natbib}
\usepackage{booktabs}
\usepackage{amsmath}
\usepackage{amsfonts}
\usepackage{nicefrac}
\usepackage{microtype}
\usepackage{comment}
\usepackage{fancyvrb}
\usepackage{listings}
\usepackage{algorithm}
\usepackage{algorithmicx}
\usepackage{algpseudocode}
\usepackage{subcaption}
\usepackage{cancel}
\usepackage{xspace}
\usepackage{makecell}
\usepackage{multirow}
\usepackage{blindtext}
\usepackage{multicol}
\usepackage{fontawesome5}
\usepackage{graphicx}
\usepackage{amssymb}
\usepackage{xcolor}
\definecolor{stargold}{RGB}{245,180,0}

\usepackage{tcolorbox}
\tcbuselibrary{skins}

\usepackage{pdflscape}
\usepackage{booktabs}
\usepackage{multirow}
\usepackage{array}
\usepackage{url}

\newcolumntype{L}{>{\raggedright\arraybackslash\sloppy}p{6.8cm}}
\newcolumntype{C}{>{\centering\arraybackslash}p{1.45cm}}
\usepackage{newtxtext,newtxmath}
\usepackage[scaled=0.94]{helvet}

\usepackage[colorinlistoftodos]{todonotes}

\usepackage{hyperref}
\hypersetup{
    pdfborderstyle={/S/S/W 1},
    linkbordercolor={0.4 0.7 1},
    citecolor=citeblue,
    breaklinks=true,
    colorlinks=true,
    linkcolor=red,
    urlcolor=linkblue,
}

\usepackage[colorinlistoftodos]{todonotes}

\newtcolorbox{MyBox}[1]{
    colback=white,
    colframe=black,
    sharp corners,
    breakable,
    left=2mm,
    right=2mm,
    top=2mm,
    bottom=2mm,
    boxrule=0.1mm,
    fontupper=\ttfamily\small
}
\definecolor{GLM_bg}{RGB}{235, 245, 255}
\definecolor{GLM_text}{RGB}{0, 110, 220}
\definecolor{header_gray}{RGB}{242, 242, 242}
\definecolor{section_gray}{RGB}{230, 230, 230}

\definecolor{neg}{HTML}{CB4335}
\definecolor{pos}{HTML}{27AE60}
\definecolor{delta_gray}{HTML}{6C8EBF}
\definecolor{light_bg}{RGB}{235, 245, 255}

\hypersetup{
  colorlinks=true,
  urlcolor=linkblue,
  linkcolor=linkblue,
  citecolor=linkblue
}

\newtcolorbox{abstractpanel}{
  enhanced,
  colback=abstractpurple!3,
  colframe=abstractpurple!45,
  boxrule=0.4pt,
  arc=2mm,
  outer arc=2mm,
  left=6mm,
  right=6mm,
  top=3mm,
  bottom=3mm,
  before skip=5mm,
  after skip=4mm,
  fontupper=\fontsize{10}{12}\selectfont
}

\begin{document}
\thispagestyle{firststyle}

\begingroup
\setlength{\parindent}{0pt}

\vspace{3mm}

{\centering\normalfont\fontfamily{ppl}\bfseries\color{black}
    {\fontsize{15}{22}\selectfont Qwen-Planner-Agent: A Closed-Loop AI-for-AI Framework for Real-World Mobile Planner Agents\par}
}

\vspace{3mm}

{\centering\normalfont\normalsize
MAI Team, Alibaba Token Hub, Alibaba Group\par}

\vspace{2mm}

\vspace{0.3em}
{\centering\normalfont\footnotesize
\hypersetup{urlcolor=magenta}
\urlstyle{tt}
\faGithub\enspace\url{https://tongyi-mai.github.io/Qwen-Planner-Agent/}\par}

\vspace{3mm}

\begin{abstractpanel}
{\centering\sffamily\bfseries\fontsize{12}{12}\selectfont\textcolor{abstractpurple!80!black}{Abstract}\par}
\vspace{2mm}

%
The rapid progression of large language models is extending AI from passive content generation into the active workflows of engineering and scientific discovery. This shift raises a compelling question: can AI be both the object of development and an active participant in building next-generation AI systems? We explore this question by building \textbf{Qwen-Planner-Agent} within a closed-loop AI-for-AI framework for scalable development and iterative improvement. Mobile planning offers a demanding test of this approach: complex, long-horizon tasks challenge agent reliability, while costly real-device interaction limits development scalability. The framework connects data production, model training, and deployment through a shared action-feedback-verification contract. (i) \emph{AI for Data} builds a human-gated agentic data flywheel in which specialized agents construct tasks, collect interaction trajectories, curate and balance training data, and use training feedback to guide subsequent data generation. (ii) \emph{AI for Training} combines a supervised planning cold start with hybrid-environment online agentic reinforcement learning, where we introduce Competence-Aware Reward-and-Advantage Engineering (CARE) to reduce reasoning and tool-use costs while preserving task performance. (iii) \emph{AI} drives model--harness co-evolution through an execution-evidence-driven loop that orchestrates memory, skills, and tools at runtime and feeds structured action feedback and preserved failure traces back into coordinated model and harness adaptation. Qwen-Planner-Agent achieves the best overall performance among all evaluated models and systems on MobilePA-Bench, improving over its base model across tool use, memory, skills, and sub-agent coordination. Further evaluations of our model show improvements across non-mobile agentic benchmarks while largely preserving general capabilities. 




\end{abstractpanel}
\endgroup

\noindent
\begin{minipage}{\linewidth}
  \centering
  \includegraphics[
    width=0.99\linewidth,
    height=\textheight,
    keepaspectratio
  ]{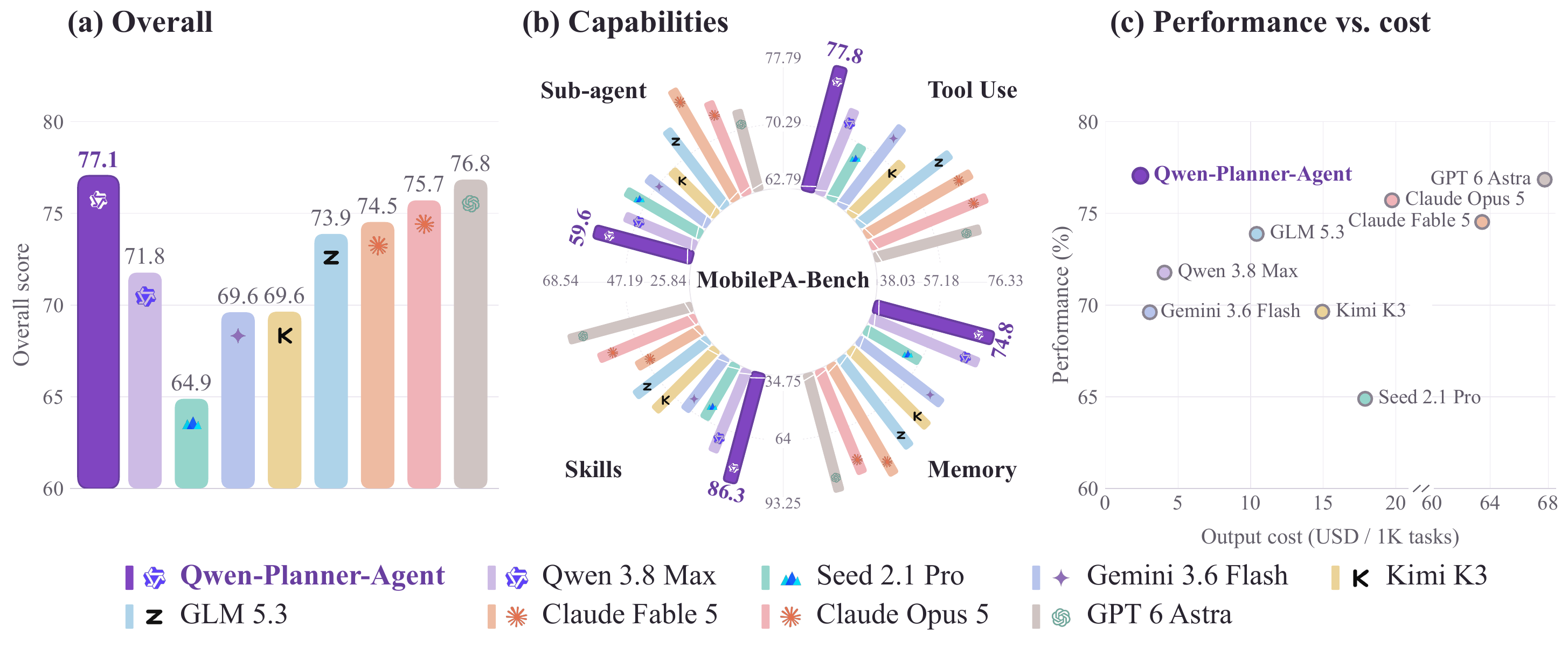}
  \captionof{figure}{Qwen-Planner-Agent achieves frontier-level Overall performance on MobilePA-Bench (left), with strong tool-use, memory, skill-use, and sub-agent capabilities (center) at a low estimated output cost including thinking tokens (right).}
  \label{fig:performance}
\end{minipage}

\clearpage

\clearpage
\begingroup
\hypersetup{linkcolor=black}
\tableofcontents
\endgroup
\clearpage

\section{Introduction}


\begin{tcolorbox}[
  enhanced,
  colback=abstractpurple!8!white,
  coltext=black!85,
  frame hidden,
  boxrule=0pt,
  borderline west={2pt}{0pt}{abstractpurple!75!white},
  sharp corners,
  left=4mm,
  right=4mm,
  top=3mm,
  bottom=3mm,
  before skip=2mm,
  after skip=4mm,
  fontupper=\small
]
\noindent\emph{``Since the design of machines is one of these intellectual activities, an ultraintelligent machine could design even better machines; \ldots''}
\par\smallskip
{\raggedleft\footnotesize\color{black!65}--- I.\ J.\ Good, \textit{Speculations Concerning the First Ultraintelligent Machine}, 1965~\citep{good1965speculations}.\par}
\end{tcolorbox}

Recent advances in language-model agents are extending AI beyond generating individual outputs toward executing multi-step workflows guided by interaction and feedback~\citep{wang2023voyager}. These advances make it increasingly practical to pursue a long-standing ambition: using AI to help build and improve AI. As early as 1950, Turing proposed developing machine intelligence by educating a ``child machine'' and iteratively refining its design through experimentation~\citep{turing1950computing}. Today, language-model agents invite a further step: involving AI not only as the system being developed, but also as a participant in the development process. This raises a central question: how can these capabilities be organized into a scalable, feedback-driven development lifecycle? We study \emph{AI for AI} from this perspective, using execution experience to guide coordinated improvements in data production, model training, and deployment.


We investigate this question by developing a mobile planner agent for complex, real-world tasks.
Completing a high-level user goal requires coordinating actions across applications, maintaining context as states change, recovering from failures, and verifying that the intended outcome has been reached.
Developing these capabilities calls for diverse interaction data, effective policy learning, and runtime support suited to changing tasks and resources.
Yet real-device interaction is costly and difficult to parallelize, constraining the scale of development and evaluation~\citep{bai2024digirl,tang2026phonebuddy}.
Task-specific verification provides a complementary opportunity: execution traces and observable outcomes can supply evidence for assessing progress and guiding targeted improvements.
Mobile planning therefore provides a concrete setting for studying scalable, AI-assisted agent development.


\begin{samepage}
In this report, we present an \emph{AI-for-AI} framework for developing \textbf{Qwen-Planner-Agent}, a complete agent system that couples a trained Planner Model with a unified Harness. Within this framework, AI interprets execution feedback to identify capability gaps and guide coordinated updates to training data, learning strategies, and runtime support.
The \emph{AI for Data} stage combines AI-assisted task construction and failure diagnosis with automated trajectory collection and curation. Training and development-set feedback guides task generation and sampling adjustments.
The \emph{AI for Training} stage combines a planning-oriented cold start with hybrid-environment online agentic RL. Competence-Aware Reward-and-Advantage Engineering (CARE) adapts rewards and calibrates advantages, with a bounded LLM-based controller configuring predefined reward schedules from training and validation feedback.
The \emph{AI for Harness} stage integrates the Planner Model with a Harness that supplies tool-conditioned Skills, persistent Memory, and execution feedback. AI assists memory consolidation, while failure diagnosis guides data updates and LLM-based Harness revisions.
\end{samepage}

The development process adapts as the agent's capabilities evolve. Diagnosed failures guide new tasks and data sampling, while updated model behavior informs Harness revisions. In turn, revised Harness instructions shape the context and interaction experience used in subsequent model learning. This reciprocal adaptation links model improvement with runtime refinement across development rounds. Model parameters remain fixed during serving; offline updates are validated and versioned, with human review retained for ambiguous and release-critical decisions.

On MobilePA-Bench, Qwen-Planner-Agent 27B achieves the highest Overall score among the evaluated models and agent systems. Its estimated per-task output cost, including thinking tokens, is lower than that of the commercial LLMs in our cost comparison. Beyond mobile planning, our Planner Model, Qwen-Planner-Model, demonstrates broad planning and tool-use capabilities across general agentic benchmarks. Ablation studies further support the effectiveness of model--Harness co-evolution in both mobile planning and general agentic settings.

\Needspace{5\baselineskip}
In summary, our contributions are threefold:
\begin{itemize}
    \item \textbf{A closed-loop AI-for-AI framework.}
    We present a practical exploration of AI-for-AI through the development of mobile planner agents. AI turns execution feedback into targeted improvements in training data, learning strategies, and runtime support. By adapting these development decisions to evolving agent capabilities and leveraging scalable hybrid environments, our framework provides a closed-loop approach to building and iteratively improving real-world agents.

    \item \textbf{Qwen-Planner-Agent.}
    We develop a unified Model--Harness agent system that couples generalizable planning with adaptive runtime support. The Planner Model learns task decomposition, grounded tool use, and failure recovery, while the Harness assembles tool-conditioned Skills, persistent Memory, and execution feedback to accommodate changing resources and user context. AI consolidates execution experience and diagnoses failures to guide model training and Harness refinement. These reviewed updates provide a pathway toward model--Harness co-evolution.

    \item \textbf{Performance, generalization, and efficiency.}
    Qwen-Planner-Agent achieves the highest Overall score among the evaluated frontier models and agent systems on MobilePA-Bench, demonstrating strong mobile planning and task-completion capabilities at a lower estimated per-task output cost than the commercial LLMs included in our cost comparison. The Planner Model also demonstrates broad competence across general agentic benchmarks, showing that its planning and tool-use capabilities extend beyond mobile environments.
\end{itemize}

\Needspace{8\baselineskip}
\section{Qwen-Planner-Agent}
\label{qobile}

\subsection{Overview}
\label{sec:system-overview}

\subsubsection{AI-for-AI Framework}

\begin{figure}[!t]
    \centering
    \includegraphics[width=0.99\linewidth]{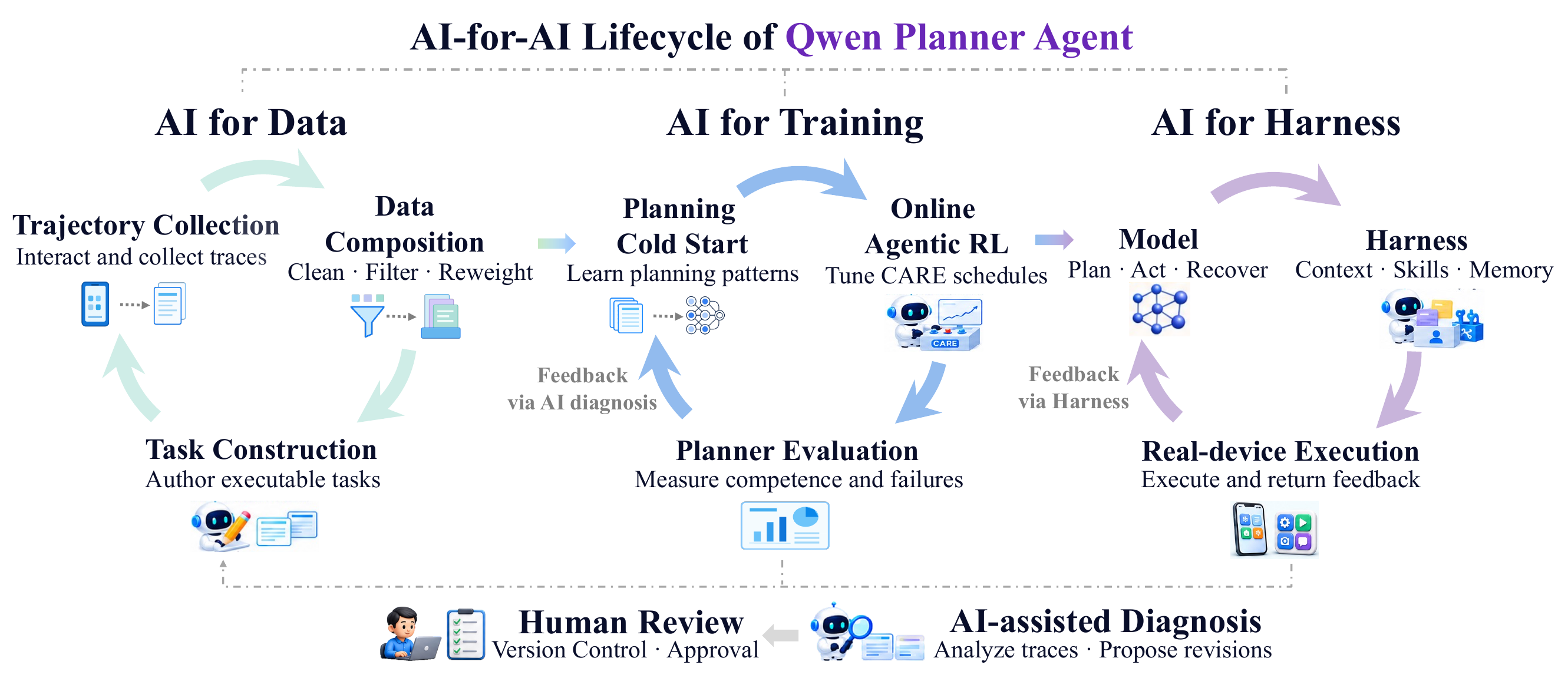}
    \caption{AI-for-AI lifecycle of Qwen-Planner-Agent, comprising three interconnected phases. (i) AI for Data combines AI-assisted task generation with automated trajectory collection and curation, using model feedback to refine subsequent tasks and training-data composition. (ii) AI for Training combines planning-oriented cold-start training with hybrid-environment online reinforcement learning, while CARE adapts reward scheduling and advantage signals to the model’s evolving competence. (iii) AI for Harness equips the planner with a unified Harness that manages Skills, persistent Memory, and execution feedback, while AI-assisted diagnosis guides subsequent model and Harness refinement and feeds new requirements back into data production and training.}
    \label{fig:framework}
\end{figure}

Figure~\ref{fig:framework} summarizes the AI-for-AI lifecycle for developing Qwen-Planner-Agent, which comprises a Planner and a Harness. \emph{AI for Data} (Section~\ref{sec:data}) combines AI-assisted task construction with automated interaction collection and curation to supply training tasks and trajectories. \emph{AI for Training} (Section~\ref{sec:rl}) learns the planner from these assets through a planning-oriented cold start and competence-adaptive online agentic RL. \emph{AI for Harness} (Section~\ref{sec:harness-engineering}) equips the planner with a Harness for Skills, Memory, and execution feedback, with AI assisting memory consolidation and Harness refinement.

Within the AI-for-AI lifecycle, intermediate evaluation uses a held-out development set rather than the final benchmark test set. Development tasks and their trajectories are excluded from direct training, while their evaluation results and diagnosed failure patterns guide subsequent data generation, training adjustments, and Harness refinement.

Training and development-set results, together with deployment traces, feed AI-assisted diagnosis that guides targeted tasks, data reweighting, and Harness revisions. Revised Harness instructions shape the context and trajectories used for subsequent model training, while updated model behavior informs further Harness refinement. This feedback connects the three stages across development rounds. Model and Harness updates are reviewed and versioned offline; model parameters remain fixed during serving.

\subsubsection{Task Formulation}
\label{sec:task-formulation}

Given a user request $x$, Qwen-Planner-Agent interacts with a partially observed environment initialized at state $s_0$. The underlying state $s_t$ includes the environment and execution-relevant runtime state and is not directly exposed to the policy. Instead, the policy receives observations $o_t$ through the environment interface. At step $t$, the Harness combines the action--observation history with retrieved memory $m_t$ and loaded skills $\mathcal K_t$ to construct the model context. The policy selects an action from the currently available structured action set $\mathcal A_t$:
\begin{equation}
    c_t=\mathcal H_\eta(x,m_t,\mathcal K_t,o_{\leq t},a_{<t}),
    \quad
    a_t\sim\pi_\theta(\cdot\mid c_t,\mathcal A_t).
\end{equation}
Here $\mathcal H_\eta(\cdot)$ denotes the Harness context-construction function under instruction configuration $\eta$, $c_t$ the resulting model context, and $\pi_\theta$ the policy parameterized by $\theta$. The environment executes $a_t$, transitions to the next underlying state $s_{t+1}$, and returns the next observation $o_{t+1}$:
\begin{equation}
    s_{t+1}\sim P_{\mathrm{trans}}(\cdot\mid s_t,a_t),
    \quad
    o_{t+1}\sim P_{\mathrm{obs}}(\cdot\mid s_{t+1},a_t).
\end{equation}
$P_{\mathrm{trans}}(\cdot)$ models environment dynamics, while $P_{\mathrm{obs}}(\cdot)$ models the structured feedback exposed to the policy, such as tool results, observable state changes, or execution errors. This formulation accommodates both deterministic and stochastic backends.

A task-specific verifier evaluates completion using the interaction history and available execution evidence:
\begin{equation}
    v_{t+1}=\mathcal V(x,\xi_{t+1},\tau_{\leq t+1}),
    \quad \text{where}\;
    \tau_{\leq t+1}=(o_{\leq t+1},a_{\leq t}).
\end{equation}
The verifier $\mathcal V(\cdot)$ produces the verification outcome $v_{t+1}$ for request $x$ from the interaction history $\tau_{\leq t+1}$ and available execution evidence $\xi_{t+1}$. The evidence $\xi_{t+1}$ includes relevant initial conditions and backend state records available to the verifier; this evidence need not be exposed to the policy. If execution continues, the returned observations enter the next model context. Interaction terminates when the verifier confirms task completion, the agent requests clarification or refuses an unsafe request, or the execution budget is exhausted. The action space covers typed tool calls, memory access and updates, skill selection and loading, clarification or refusal, and task-completion declarations. The agent primarily acts through structured tools rather than pixel-coordinate GUI actions, although tools may return visual observations when needed. Backends may differ in their internal state representations and execution mechanisms while exposing compatible task, action, observation, and verification records.

\subsection{Hybrid Mobile Environment Infrastructure}
\label{sec:hybrid-mobile-environment}

Mobile-agent development requires scalable interaction as well as feedback that reflects real execution. We combine programmatic sandboxes, LLM-simulated environments, and selected real-device sessions to support the task interactions formulated in Section~\ref{sec:task-formulation}. The infrastructure organizes these backends into shared data-collection, evaluation, and online-training workflows.

\subsubsection{Complementary Environment Backends}
\label{sec:environment-backends}

The three backends trade off scalability, scenario coverage, and execution fidelity, making each suitable for different task requirements.

\textbf{Programmatic sandbox environments} execute typed tool calls through predefined program logic over structured application databases. Deterministic transitions, task-specific resets, and state-based verification support reproducible, high-throughput interaction. Their coverage is limited to implemented tools and state transitions, so new applications or exceptional behavior require additional engineering. They are therefore most suitable for repeatable tasks with explicit state and completion conditions.

\textbf{LLM-simulated environments} use language models to generate environment responses for long-tail interactions that are difficult to implement with fixed logic. They broaden scenario coverage without requiring a dedicated programmatic implementation for every case. However, responses may be inconsistent with prior actions or environment state, so trajectories require task-specific validation before use.

\textbf{Real-device environments} execute actions in live device sessions, capturing the effects of OS permissions, authenticated services, cross-application dependencies, and runtime changes. They provide direct evidence of behavior that simulation may miss, but incur higher interaction costs, limited parallelism, and difficult resets. They are therefore valuable for tasks whose completion depends on actual device or service behavior.

\subsubsection{Hybrid Environment Strategy}
\label{sec:hybrid-environment-strategy}
\label{sec:environment-interface}

We match tasks to backends according to their execution requirements. Tasks with well-defined, reproducible state changes primarily use programmatic sandboxes, while LLM-simulated environments supplement long-tail interactions without suitable fixed implementations. Real-device sessions are used selectively for tasks dependent on live device or service behavior. This allocation combines scalable simulated interaction with prioritized device access where simulation lacks sufficient execution fidelity.

A common agent-facing interface makes these experiences usable within the same data and training workflows. Tasks expose typed actions, structured responses, and task-specific completion criteria; their interaction records retain execution errors, observable state changes, and verification outcomes. Backend implementations and internal states remain distinct. Compatible records enter shared curation and rollout-processing pipelines, without assuming that simulated and real-device feedback have identical reliability.

\subsubsection{Training Infrastructure}
\label{sec:environment-management}
Our infrastructure separates model-side training from environment-side execution. The training layer coordinates distributed rollout generation, model updates, and training-resource scheduling. A client--server environment-management layer creates, schedules, and cleans up environment instances and real-device sessions, while the corresponding backends implement resets and state handling. This separation allows model computation and environment capacity to be managed independently.

During an online rollout, a policy worker interacts with a backend through the environment-management layer and receives execution feedback after each action. Task verifiers assess completion, while environment validation and trajectory checks determine which records are admitted to training. The training layer uses the admitted trajectories for policy updates. Task failure is distinct from an invalid execution record: unsuccessful interactions can still provide learning and diagnostic evidence. Detailed data-curation rules are described in Section~\ref{sec:data}.

The environment-management layer also supports data collection and evaluation. It remains distinct from the deployment-time Harness, which assembles the Planner Model's context from Skills, Memory, tools, and execution feedback. Further training and environment-management details are provided in Appendix~\ref{app:training-details}.

\subsection{AI for Data: An Agent-Driven Data Flywheel}
\label{sec:data}

\begin{figure}[!t]
    \centering
    \includegraphics[width=\linewidth]{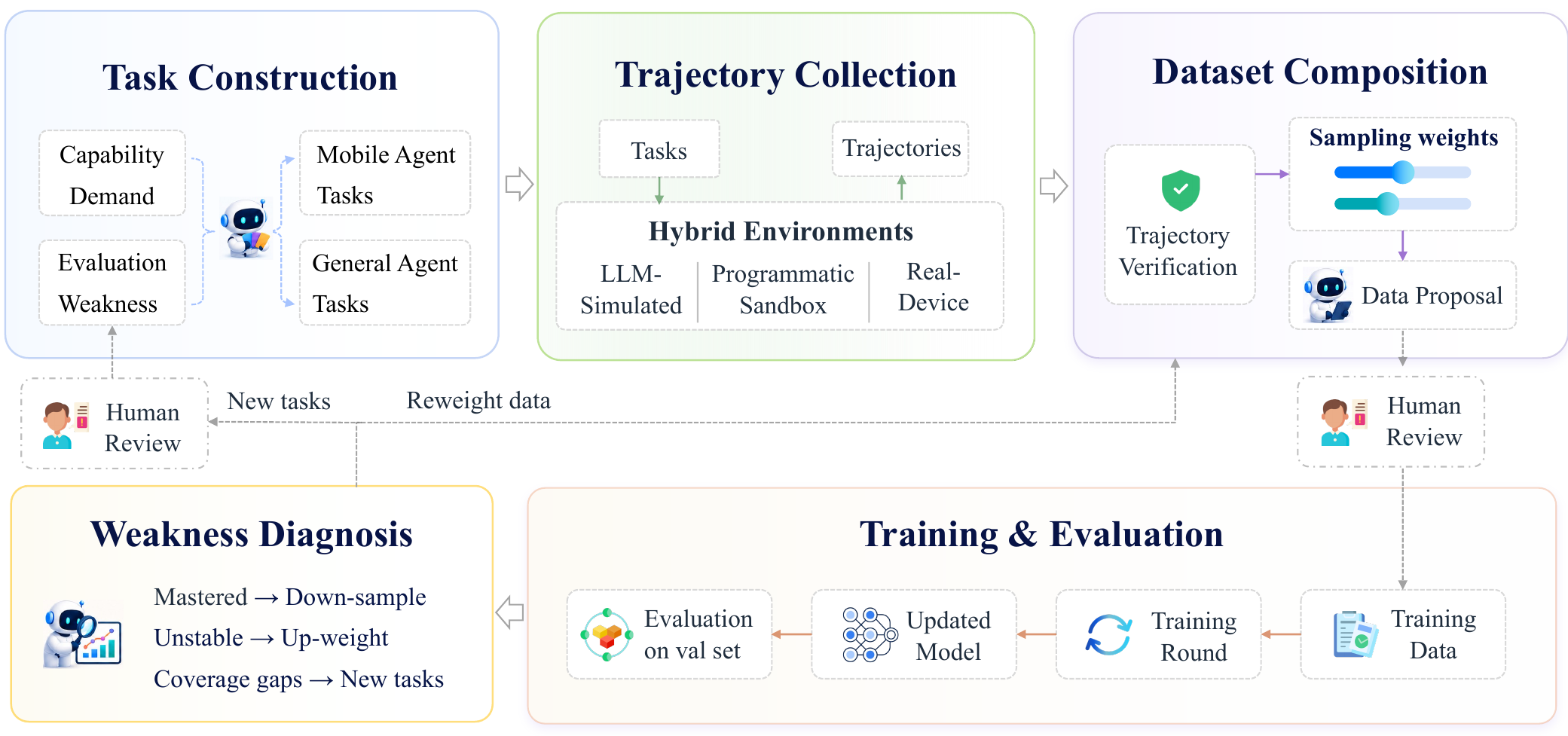}
    \caption{\textbf{AI-for-Data workflow.} AI-assisted task construction supplies executable tasks for automated interaction collection and curation. Verified trajectories and resettable tasks support planning-oriented cold-start training and online RL, respectively. AI-assisted analysis of rollout and development-set feedback guides targeted task generation and sampling adjustments for the next iteration, with human review before each data release. Solid blue arrows denote automated within-iteration flow; dashed orange arrows denote the human-gated transition between iterations.}
    \label{fig:ai_for_data_pipeline}
\end{figure}

The data flywheel in Figure~\ref{fig:ai_for_data_pipeline} turns capability requirements into executable tasks, collects interaction trajectories, and constructs training datasets that evolve with planner performance. AI assists task construction, failure diagnosis, and targeted data refinement, while automated workflows handle rollout collection and data processing. Training and development-set feedback informs which tasks to generate and how to adjust sampling in the next iteration. The resulting assets support both planning-oriented cold-start training and online reinforcement learning.

\subsubsection{Task Construction}
\label{sec:data-task-construction}

Task-construction agents translate target capabilities into executable task specifications. Initial objectives come from product requirements, available tool inventories, and representative user scenarios; later iterations also incorporate diagnosed capability gaps, as described in Section~\ref{sec:data-iteration}. Each specification contains a user goal, available resources, relevant initial conditions, target capabilities, and completion criteria. It defines what must be accomplished without prescribing a single reference trajectory, allowing different valid plans to satisfy the same objective.

Construction jointly considers \emph{scenario coverage} and \emph{capability coverage}. Scenario coverage spans application domains, tools, user intents, and interaction patterns. Capability coverage targets information acquisition, tool routing, argument grounding, multi-step dependency handling, state tracking, recovery, and verified task completion. This distinction helps introduce new behavioral requirements rather than merely adding more instances of familiar scenarios.

\subsubsection{Interaction Trajectory Collection}
\label{sec:trajectory-generation}

Tasks are executed in backends matched to their interaction requirements. Mobile tasks follow the hybrid strategy in Section~\ref{sec:hybrid-environment-strategy}; general-agent tasks use executable service environments, while coding tasks use repository and code-execution environments.

An automated rollout workflow runs agent policies through multi-step environment interaction and records their trajectories. Each record preserves selected actions, environment feedback, intermediate outcomes, execution errors, retries, recovery attempts, completion evidence, and the final outcome. Successful trajectories provide candidate supervision for planning, tool use, and task closure. Failed and incomplete trajectories are also retained so that diagnosis can examine where execution diverged, whether recovery was attempted, and why the task remained unfinished. Collection therefore preserves the process leading to an outcome, not only its final success label.

\subsubsection{Training Dataset Composition}
\label{sec:training-dataset-composition}

Task-completion verification and data-quality assessment are separate decisions. Environment verifiers determine whether completion criteria have been met, while the curation workflow checks whether the corresponding record is suitable for learning. Automated processing normalizes heterogeneous trajectories, removes malformed or duplicate examples, and checks schema consistency and support for reported outcomes in the execution evidence. Retained records are tagged by capability and linked to failure attributions from rollout analysis. Successful completion alone does not establish training suitability, and unsuccessful interactions may still provide useful diagnostic evidence.

Low-confidence, conflicting, safety-sensitive, or insufficiently supported cases are routed to human reviewers, who may approve, correct, reject, or quarantine them. Retained records preserve their task source, execution backend, generating policy, capability annotations, outcome, and failure attribution, making later sampling and repair decisions traceable.

The curated data are organized into mobile and non-mobile groups. Mobile data form the core of training and cover cross-application planning, tool execution, state tracking, Memory, Skills, sub-agent coordination, recovery, and task completion. Non-mobile data include general-agent trajectories, coding tasks, and reasoning and instruction-following examples. General-agent trajectories exercise structured tool use and multi-turn planning over services such as Model Context Protocol (MCP) servers~\citep{anthropic2024mcp}; coding tasks add repository understanding, iterative editing, and testing. These examples provide complementary supervision for multi-step problem solving while helping preserve general reasoning and instruction following.

The pipeline maintains two training assets: verified interaction trajectories for planning-oriented cold-start training and resettable task instances with reliable completion criteria for online reinforcement learning. These support learning from recorded demonstrations and newly generated interactions, respectively. We configure sampling weights across mobile and non-mobile data, capabilities, difficulty levels, and sources rather than sampling in proportion to raw corpus size. These weights are revised using the feedback described below, keeping mobile planning as the primary objective while retaining complementary non-mobile supervision.

\subsubsection{Feedback-Driven Data Refinement}
\label{sec:data-iteration}

As the planner improves, the value of individual tasks changes: mastered tasks may become redundant, whereas unstable behaviors and uncovered capabilities require additional training. AI-assisted analysis of task-level rollouts and capability-level development-set results informs three types of updates: reducing redundant examples, increasing coverage of unstable behaviors, and constructing tasks for missing capabilities.

At the task level, rollout-analysis agents examine completion outcomes and failure patterns. Reliably mastered tasks are down-sampled while retaining a small preservation set, and inconsistently completed tasks receive greater weight as stabilization data. Failures involving tool routing, argument grounding, state tracking, recovery, or premature termination are grouped into targeted repair sets. Ambiguous or weakly supported cases return to curation rather than entering the training pool as ordinary examples.

At the capability level, analysis agents combine held-out development-set results for Tool Use, Memory, Skills, and Sub-agent with trace-derived failure distributions to distinguish recurring gaps from isolated errors. If the existing pool contains suitable examples, their sampling proportions are adjusted. If coverage is insufficient, task-construction agents generate new tasks with targeted capability requirements, difficulty levels, tool combinations, or interaction patterns. Feedback thus determines both which data to select and which tasks to construct next. Development-set failure patterns guide the construction of new training tasks; the development tasks and their trajectories are not added to the training pool.

Proposed task additions and removals, sampling adjustments, repair and quarantine decisions, and quality indicators are recorded together for review. Human reviewers may approve, revise, or reject these changes; only an approved dataset and sampling configuration are frozen for the next training round. Each release preserves a versioned history linking diagnosed capability gaps to the data updates made in response. The feedback loop thus changes both the content and composition of subsequent training data, rather than simply expanding the corpus.

\subsection{AI for Training: Competence-Adaptive Agentic Optimization}
\label{sec:rl}

\begin{figure}[!t]
    \centering
    \includegraphics[width=\linewidth]{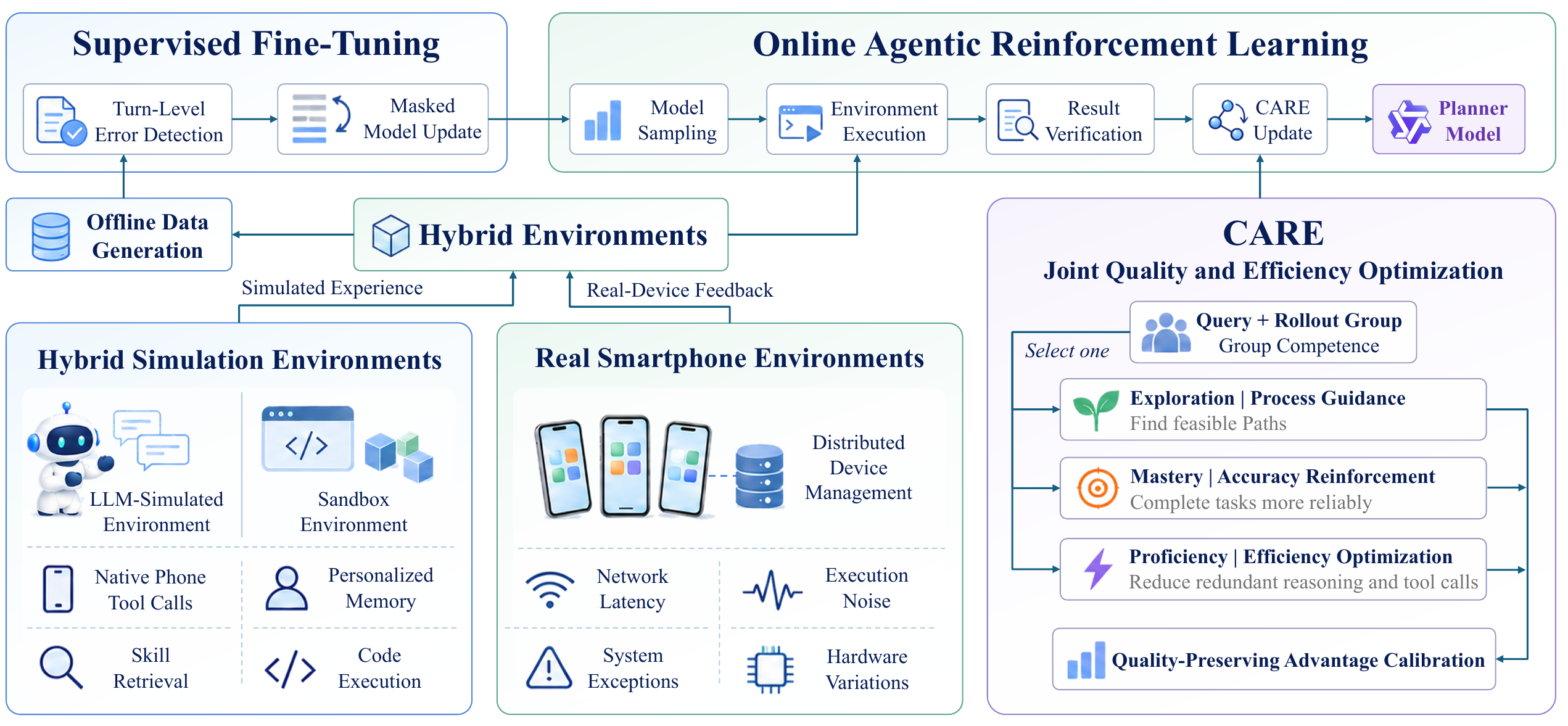}
    \caption{\textbf{Training framework for the Planner Model in Qwen-Planner-Agent.} Supervised fine-tuning with turn-level error detection and loss masking establishes a planning-oriented cold-start policy, followed by online agentic reinforcement learning in hybrid environments combining LLM simulation, programmatic sandboxes, and selected real-device interaction. Within the RL stage, CARE (Competence-Aware Reward-and-Advantage Engineering) selects process guidance, accuracy reinforcement, or efficiency optimization according to each trajectory group's competence. Quality-preserving advantage calibration limits the amplification of small efficiency differences, promoting efficient reasoning and tool use while retaining verified task success as the primary objective.}
\label{fig:rl-framework}
\end{figure}

The training pipeline begins with a planning-oriented cold start that converts curated data from the AI-for-Data pipeline into an initial policy following the shared action--feedback--verification contract. Starting from this policy, we perform hybrid-environment online agentic reinforcement learning across programmatic sandboxes, LLM-simulated environments, and selected real-device sessions. Within the RL stage, we propose Competence-Aware Reward-and-Advantage Engineering (CARE), which adapts reward composition and advantage scaling to measured group competence to improve reasoning and tool-use efficiency while preserving task quality. Figure~\ref{fig:rl-framework} summarizes this process.

\subsubsection{Planning-Oriented Cold Start}

We initialize the policy through supervised fine-tuning on curated data from the AI-for-Data pipeline, learning task decomposition, action sequencing, tool use, state-conditioned replanning, and recovery under the shared action--feedback--verification contract. Training prioritizes verified mobile trajectories and includes non-mobile data for broader agentic capabilities, general reasoning, and instruction following, with controlled sampling weights across data groups. After sample-level filtering, we mask model-action turns identified as erroneous during trajectory curation. For each trajectory $\tau_i$ in a training batch, let $\mathcal E_i$ denote the set of erroneous turn indices and $\boldsymbol y_{i,j}$ the token sequence generated in turn $j$. We minimize the masked cross-entropy objective
\begin{equation}
    \mathcal L_{\mathrm{SFT}}(\theta)
    = -Z^{-1}\sum_{i,j}\mathbf{1}\{j\notin\mathcal E_i\}
    \log\pi_\theta(\boldsymbol y_{i,j}\mid h_{i,j}),
\end{equation}
where $h_{i,j}$ is the conditioning context preceding turn $j$ and $Z$ is the total number of unmasked model-generated tokens in the batch. Execution feedback is retained in the context, while verified recovery turns remain supervised.

\subsubsection{Hybrid-Environment Online Agentic Reinforcement Learning}
\label{sec:hybrid-rollout-strategy}

Starting from the cold-start policy, we use online agentic reinforcement learning to strengthen long-horizon planning, state tracking, and failure recovery. Curated demonstrations establish useful execution patterns but cannot cover all interaction states and failure modes encountered as the policy evolves. Online rollouts expose the planner to this policy-dependent distribution, with verifier feedback guiding improvements in task completion and robustness.

\subsubsection{CARE: Competence-Aware Reward-and-Advantage Engineering}

CARE adapts reward composition and advantage scaling to group competence to promote reliable task completion and efficient execution. Its central principle is to emphasize reasoning and tool-use efficiency as task success becomes reliable, while retaining verified success as the primary learning objective. A fixed reward design is poorly matched to groups at different competence levels: groups with sparse success benefit from verified progress signals, groups with mixed outcomes should consolidate task completion, and groups whose success has saturated can shift attention toward execution efficiency. Moreover, standard within-group normalization can over-amplify small efficiency differences once success has saturated. CARE addresses these two issues through competence-adaptive reward scheduling and quality-preserving advantage calibration. A bounded LLM-based controller periodically configures the predefined scheduling parameters from recent training statistics and held-out development-set feedback.

For each task, we sample a group of $G$ trajectories, $\mathcal{G}=\{\tau_i\}_{i=1}^{G}$ using a rollout behavior policy. A task-specific rule-based verifier or rubric-guided generative reward model assigns each trajectory a binary success indicator $s_i\in\{0,1\}$. The group success rate $\overline s=\frac{1}{G}\sum_i s_i$ serves as the competence signal for reward scheduling and advantage calibration.

\paragraph{Competence-Adaptive Reward Scheduling}
CARE defines three competence-dependent reward regimes: \emph{progress shaping}, \emph{outcome consolidation}, and \emph{efficiency refinement}. Each trajectory group is assigned to a regime according to its current success rate $\overline s$, allowing different regimes to coexist within the same training batch.
\begin{equation}
    R_i = s_i +
    \left\{
    \begin{array}{cll}
        \lambda_{\mathrm{prog}}R_{\mathrm{prog},i},
        & \overline s\in[0,p_{\mathrm{low}}),
        & \text{progress shaping} \\
        0,
        & \overline s\in[p_{\mathrm{low}},p_{\mathrm{high}}),
        & \text{outcome consolidation} \\
        -\lambda_{\mathrm{eff}}e_i,
        & \overline s\in[p_{\mathrm{high}},1],
        & \text{efficiency refinement}
    \end{array}
    \right.
\end{equation}
Here $R_{\mathrm{prog},i}$ measures verified progress, and $e_i$ is a normalized execution-cost penalty. The group-success thresholds satisfy  $0 \leq p_{\mathrm{low}} <p_{\mathrm{high}} < 1$ , and the reward weights $\lambda_{\text{prog}}$ and $\lambda_{\text{eff}}$ are nonnegative.

\paragraph{AI-Assisted Schedule Configuration}
The competence regimes determine the auxiliary reward for each trajectory group, while their thresholds and reward weights can be adjusted as training evolves. An LLM-based controller configures reward-scheduling parameters $\boldsymbol{\phi}=\{\lambda_{\mathrm{prog}},\lambda_{\mathrm{eff}},p_{\mathrm{low}},p_{\mathrm{high}}\}$ every $N$ policy-update steps. Recent on-policy training statistics characterize the optimization state, while held-out development-set performance serves as the primary signal for selecting the configuration used over the next $N$ steps:
\begin{equation}
    \boldsymbol{\phi}_{t+1:t+N}
    = f_{\mathrm{LLM}}\!\left(
    \mathcal{P},
    \mathcal{S}^{\mathrm{train}}_{\leq t},
    \mathcal{M}^{\mathrm{val}}_{\leq t}
    \right),
    \quad t\in\{0,N,2N,\ldots\}
\end{equation}
Here $\mathcal{P}$ specifies the optimization objective, parameter semantics, admissible ranges, and output format. $\mathcal{S}^{\mathrm{train}}$ summarizes group success, progress, and efficiency statistics, and $\mathcal{M}^{\mathrm{val}}$ records success and execution quality on the held-out development set. Each configuration is held fixed for the subsequent $N$ policy updates and applied consistently in reward scheduling and advantage calibration; individual groups continue to select their competence regime according to their current success rate.

\paragraph{Quality-Preserving Advantage Calibration}
Reward scheduling controls reward composition, but within-group normalization can counteract the intended attenuation of efficiency-only learning signals. Standard GRPO-style~\citep{shao2024deepseekmath} advantage computation centers rewards and divides by their within-group standard deviation, bringing groups with nonzero reward variation to approximately unit advantage scale. Once success saturates, small efficiency differences can therefore produce advantages comparable in scale to those of groups with mixed success outcomes. To examine this effect, we consider the efficiency-refinement regime, where $R_i=s_i-\lambda_{\mathrm{eff}}e_i$. Let $\overline R=\frac{1}{G}\sum_jR_j$ and $\overline e=\frac{1}{G}\sum_je_j$. The standardized advantage and reward variance are
\begin{equation}
    \begin{gathered}
        \widehat A_i^{\mathrm{std}}
        = \frac{R_i-\overline R}{\sigma_R+\epsilon}
        = \frac{(s_i-\overline s)-\lambda_{\mathrm{eff}}(e_i-\overline e)}
        {\sigma_R+\epsilon} \\
        \text{where}\quad
        \sigma_R^{2}
        = \overline s(1-\overline s)
        +\lambda_{\mathrm{eff}}^{2}\sigma_e^{2}
        -2\lambda_{\mathrm{eff}}\operatorname{Cov}(s,e)
    \end{gathered}
\end{equation}
For a fully successful group ($s_i=1\;\text{for}\;\text{all}\; i$), the success variance and its covariance with $e$ are zero. For $\lambda_\text{eff}>0$ and $\sigma_e >0$, the expressions simplify to
\begin{equation}
    \begin{gathered}
        \lim_{\overline s\to1}(R_i-\overline R)
        =-\lambda_{\mathrm{eff}}(e_i-\overline e)
        ,\quad 
        \lim_{\overline s\to1}\sigma_R
        =\lambda_{\mathrm{eff}}\sigma_e \\
        \lim_{\overline s\to1}\widehat A_i^{\mathrm{std}}
        =-\frac{\lambda_{\mathrm{eff}}(e_i-\overline e)}
        {\lambda_{\mathrm{eff}}\sigma_e+\epsilon}
        \approx-\frac{e_i-\overline e}{\sigma_e}
    \end{gathered}
\end{equation}
When $\epsilon$ is negligible relative to $\lambda_{\mathrm{eff}}\sigma_e$, the efficiency coefficient approximately cancels under normalization. Even with a small efficiency weight (e.g., $\lambda_{\mathrm{eff}}=0.1$), advantages in success-saturated groups therefore remain approximately unit scale. This gives efficiency-only groups an advantage scale comparable to that of groups with mixed success outcomes, weakening the intended emphasis on task success and potentially encouraging excessive trajectory compression.

An alternative is to normalize reward components separately before aggregation, as in GDPO~\citep{gdpo}. However, scaling each component by its own within-group standard deviation changes its effective weight in the original reward space and can alter trajectory rankings under the composed reward. We instead preserve the composed reward and calibrate its group-level advantage scale by imposing a success-derived floor on the normalization denominator:
\begin{equation}
    \begin{gathered}
        \sigma_{\mathrm{anchor}}
        = \sqrt{p_{\mathrm{high}}(1-p_{\mathrm{high}})}
        \\
        \widehat A_i =
        \left\{
        \begin{array}{ll}
            \dfrac{R_i-\overline R}{\sigma_R+\epsilon},
            & \text{progress or outcome regime}, \\[6pt]
            \dfrac{R_i-\overline R}
            {\max(\sigma_R,\sigma_{\mathrm{anchor}})+\epsilon},
            & \text{efficiency-refinement regime}
        \end{array}
        \right.
    \end{gathered}
\end{equation}
Calibration uses the same efficiency-refinement threshold $p_{\mathrm{high}}$ as reward scheduling, with the anchor set to the corresponding binary-success standard deviation. For fully successful groups where the floor is active, the within-group standard deviation of the calibrated advantages is $\lambda_{\mathrm{eff}}\sigma_e/(\sigma_{\mathrm{anchor}}+\epsilon)$, preserving its dependence on the efficiency weight. The shared denominator retains the relative weighting of reward components while limiting the amplification of small efficiency differences.

Given the calibrated advantages, we maximize the following clipped group-relative surrogate objective~\citep{schulman2017ppo,yu2026dapo}:
\begin{equation}
    \mathcal{J}_{\mathrm{RL}}(\theta)
    =
    \mathbb{E}\!\left[
    \frac{1}{G}
    \sum_{i=1}^{G}
    \frac{1}{T_i}
    \sum_{t=1}^{T_i}
    \min\!\left(
        \rho_{i,t}(\theta)\widehat A_i,
        \operatorname{clip}\!\left(
            \rho_{i,t}(\theta),
            1-\epsilon_{\mathrm{low}},
            1+\epsilon_{\mathrm{high}}
        \right)\widehat A_i
    \right)
    \right]
\end{equation}
where $\theta$ and $\theta_{\mathrm{old}}$ denote the current and rollout behavior policy parameters, respectively; $y_{i,t}$ is the $t$-th generated token in trajectory $\tau_i$; $h_{i,t}$ is the interaction history preceding it; $\rho_{i,t}(\theta)=\pi_{\theta}(y_{i,t}\mid h_{i,t})/\pi_{\theta_{\mathrm{old}}}(y_{i,t}\mid h_{i,t})$; $T_i$ is the number of generated tokens in $\tau_i$; and $\epsilon_{\mathrm{low}}$ and $\epsilon_{\mathrm{high}}$ denote the lower and upper clipping ranges.

\begin{figure}[!t]
    \centering
    \includegraphics[width=0.99\linewidth]{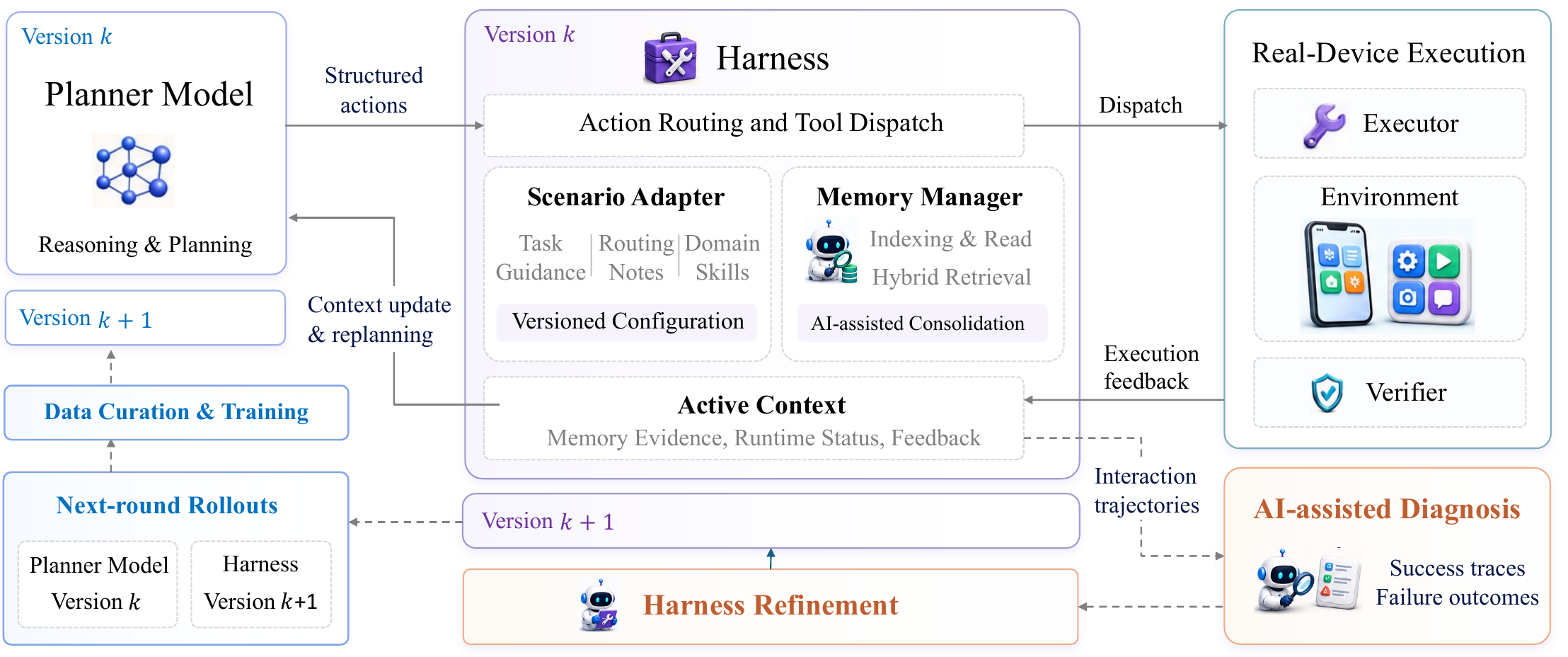}
    \caption{\textbf{Model--Harness Co-Evolution.} The Harness connects the Planner Model to deployment resources and preserves execution traces and failure evidence. Offline diagnosis of this evidence can inform subsequent data, model, and Harness refinement under a governed development process.}
    \label{fig:co_evolve}
\end{figure}

\subsection{AI for Harness: Toward Model--Harness Co-Evolution}
\label{sec:harness-engineering}

Deployment is not the endpoint of model development, but a continuing source of executable evidence. A trained Planner Model can acquire generalizable capabilities in task decomposition, tool selection, argument grounding, state tracking, and failure recovery. However, deployment-time tool inventories, operating rules, user information, and interaction histories continue to change. Encoding every change in model parameters would require repeated retraining, whereas placing all possible rules and historical records in the prompt would create long, noisy contexts.

We therefore place a unified \emph{Harness} between the Planner Model and its deployment environment. At request time, the Harness connects the model to external executors, assembles an active context from Tools, Skills, persistent Memory, and runtime constraints, and relays structured feedback after each action. Across development cycles, it preserves deployment evidence for AI-assisted trajectory analysis, failure diagnosis, and candidate refinement of both the model and the Harness. AI for Harness thus operates at two timescales: online, the Harness adapts the model context to the current request; offline, accumulated evidence informs subsequent model and Harness updates. These updates remain reviewed and versioned rather than being applied autonomously during serving.

Agent improvement involves two coupled forms of adaptation: updating the policy and adapting the execution scaffold under which it operates. We develop a model--Harness co-evolution framework that connects these channels through an iterative feedback loop. The Harness plays a dual role: it shapes the agent's current execution behavior and the experience distribution underlying subsequent model training. Conversely, model updates change the behaviors and limitations that Harness adaptation must address. Each component therefore changes the conditions under which the other is optimized.

Let $\eta$ denote the editable instruction configuration of the Harness introduced in Section~\ref{sec:system-overview}, comprising general and domain-specific instructions. The policy $\pi_\theta$ and the configured context-construction function $\mathcal H_\eta$ induce the trajectory distribution $p_{\theta,\eta}(\tau\mid x)$ through environment interaction on task $x$. We seek to maximize the joint objective
\begin{equation}
    J(\theta,\eta)
    = \mathbb{E}_{\substack{x\sim\mathcal D\\
    \tau\sim p_{\theta,\eta}(\cdot\mid x)}}
    \left[R_x(\tau)\right],
    \label{eq:coevolution_objective}
\end{equation}
where $\mathcal D$ spans heterogeneous task environments and $R_x$ represents task-dependent feedback. Resource budgets and task-specific interaction requirements define the practical operating conditions of this optimization.

Our current implementation pursues this objective through alternating model updates and feedback-driven Harness revisions:
\begin{equation}
    \begin{aligned}
        \theta^{(k+1)}
        &= \mathcal U_M(\theta^{(k)};\mathcal B^{(k)},\eta^{(k)}),\\
        \eta^{(k+1)}
        &= \mathcal U_H\!\left(\eta^{(k)};\mathcal F(\theta^{(k+1)},\eta^{(k)})\right).
    \end{aligned}
    \label{eq:coevolution_updates}
\end{equation}
Here, $k$ indexes co-evolution rounds, and $\mathcal U_M$ performs reinforcement learning using trajectories $\mathcal B^{(k)}$ generated under the fixed Harness instruction configuration $\eta^{(k)}$. The updated model is then evaluated on the held-out development set with the same configuration to obtain development-set feedback $\mathcal F(\theta^{(k+1)},\eta^{(k)})$, which an LLM-based editor $\mathcal U_H$ uses to revise the execution instructions. The revised configuration $\eta^{(k+1)}$ is applied through $\mathcal H_{\eta^{(k+1)}}$ in the next training round, closing the loop between parameter learning and contextual adaptation. This coupling makes Harness revision consequential beyond its immediate execution effects: it also reshapes the experience from which the model subsequently learns. The framework thus provides a basis for examining how changes in execution behavior translate into learning and system-level progress over successive rounds.

\subsubsection{Unified Model--Harness Runtime}
\label{sec:model-harness-runtime}

The Harness defines the runtime boundary among the Planner Model, external resources, and the execution environment. Given a user request, it assembles the active context from conversation history, currently exposed tool specifications, operational guidance supplied by the Scenario Adapter, evidence retrieved by the Persistent Memory Manager, and applicable runtime constraints. The Planner Model then produces a structured action. A separate Executor applies the action to the application or device and returns a tool result, an observable state change, or an execution error. This feedback enters the next planning turn unless completion is verified, user clarification is required, or a safety or execution-budget condition stops the interaction.

The components have distinct responsibilities. The Planner Model interprets the request, decomposes the task, resolves conflicts, selects actions, fills arguments, and replans after feedback. The Scenario Adapter conditions procedural guidance on the tools available for the current request, whereas the Persistent Memory Manager stores, governs, and retrieves evidence across interactions. Actions that modify the target application or device state are carried out by the Executor. The Harness coordinates these interfaces and retains their joint trace without taking over the Planner Model's decision-making role. This separation keeps generalizable planning behavior in the model while allowing deployment-specific information to change outside its parameters.

\subsubsection{Context Management with Skills and Memory}
\label{sec:adaptive-context-orchestration}

Skills and Memory provide complementary forms of deployment-time context. Skills describe \emph{how} available resources should be used, including tool dependencies, operating procedures, and multi-turn interaction requirements. Memory provides evidence about \emph{what} is already known or has previously occurred, including user preferences, past events, established decisions, and pending intentions. The Harness assembles tool-conditioned Skill guidance and retrieved Memory with tool specifications, interaction history, and runtime constraints. When explicit Skill-selection tools are exposed, the Planner Model selects and loads Skills through the structured action interface.

\paragraph{Scenario-conditioned Skill guidance.}
The Scenario Adapter handles variation in deployment-time tool inventories without modifying the Planner Model's parameters. Offline, it compiles tool schemas, a complete tool-to-Skill mapping, capability-domain Skills, routing notes, and guidance constraints into a versioned configuration. At request time, it treats the tools already exposed to the model as the candidate set, selects the corresponding Skill guidance, removes rules that depend exclusively on unavailable resources, and adds applicable routing and multi-turn instructions. The resulting guidance is incorporated into the active context. The Scenario Adapter neither discovers additional tools nor executes actions; it provides procedural knowledge matched to resources already available to the Planner Model.

\paragraph{Evidence-grounded Persistent Memory.}
The Persistent Memory Manager handles information that must remain available beyond the active conversation context. Persistent state is organized by function rather than stored as an undifferentiated interaction history. A compact user model records stable preferences and behavioral requirements; episodic Memory preserves detailed observations and dialogue evidence; curated long-term Memory stores consolidated facts, decisions, and reusable lessons; and prospective Memory represents future intentions with explicit activation conditions, scope, expiry, and completion status. These Memory types have different access and trust boundaries, preventing all historical content from being treated as equally reliable or relevant.

Memory follows a controlled lifecycle of \emph{capture}, \emph{validation}, \emph{indexing}, \emph{retrieval}, \emph{consolidation}, and \emph{revision}. Each Memory unit links a concise structured summary to source metadata and original evidence, supporting efficient retrieval without sacrificing traceability. Bounded AI-assisted processing helps summarize long trajectories, merge duplicate observations, identify contradictions, and formulate candidate updates. These candidates do not automatically become authoritative Memory: provenance, trust, recency, and conflict checks precede consolidation. When new evidence changes an existing fact, the revised entry explicitly supersedes the obsolete record rather than leaving competing versions unresolved.

At inference time, the Persistent Memory Manager retrieves a compact evidence set for the current task. Retrieval combines semantic and keyword matching with relevance, recency, importance, and diversity controls. Evidence assembly also reflects the structure of the request: temporal questions require event ordering, cross-session aggregation requires entity and action deduplication, and questions about changing facts require explicit resolution of superseding evidence. Retrieved Memory is supplied to the Planner Model as evidence rather than as an executable instruction, and it does not acquire permission to invoke tools or alter external state.

The Scenario Adapter and Persistent Memory Manager therefore serve the same context-orchestration process rather than forming separate runtime systems. The former contributes procedural guidance matched to the current resources, while the latter contributes historical evidence matched to the current task. Together, they preserve a shared Planner Model across heterogeneous deployment scenarios while allowing dynamic information to be updated outside model parameters. In the experiments, we distinguish only between configurations with and without the unified Harness; persistent Memory remains an internal capability of that Harness, not a separately named runtime system. Further implementation and evaluation details are provided in Appendix~\ref{app:harness-details}.

\subsubsection{Privacy-Preserving Memory Safety}
\label{sec:memory-safty}

Protecting user privacy requires security controls across the entire memory lifecycle, including collection, extraction, storage, retrieval, sharing, and deletion. The system separates private user memories from shared agent experiences. User profiles, preferences, and personal facts are restricted to the corresponding workspace, user, and agent, while reusable experiences are shared only within the designated workspace and agent. Before entering shared memory, these experiences undergo processing to reduce user-specific and sensitive details. Each extracted memory is linked to its source events for traceability, and raw processing data is cleared after task completion to limit unnecessary retention. Context files are generated from stored memories, with version checks and coordinated updates helping prevent outdated or deleted information from being reused.

Access control determines who can read, create, modify, delete, or share memories. The current implementation uses service-level authentication tokens and separate administrator credentials, while user-level access enforcement depends on trusted upstream services. For multi-user deployment, authorization should additionally verify the caller’s identity, the owner of the requested memory, and the permitted operation. Role-based access control and user-specific permission rules should enforce least privilege, with access denied unless explicitly authorized. These checks should apply to both direct memory access and semantic retrieval, ensuring that search results contain only information the requesting user is allowed to access.

Users should have clear control over what is remembered, how long it is retained, and whether it can be reused across sessions or contribute to shared experiences. The system should support memory inspection, correction, deletion, and withdrawal of permission, with changes reflected in stored records, generated context files, and relevant caches. Removing personal details from shared experiences reduces privacy risks but does not guarantee anonymity. Additional safeguards, including sensitive-information detection, encryption during transmission and storage, protected audit logs, and explicit backup retention policies, should support accountable memory use while minimizing the collection and retention of personal information.

\subsubsection{Model and Harness Refinement from Execution Feedback}
\label{sec:deployment-evidence}

Each runtime interaction produces a structured deployment trace that links the active Tool and Skill configuration, retrieved Memory and its provenance, the action and arguments selected by the Planner Model, the Executor response, observable state evidence, execution errors, and the final Verifier outcome. Preserving these relationships makes it possible to inspect how both model decisions and Harness configuration contribute to task outcomes rather than retaining only the final response.

The accumulated deployment traces and development-set feedback support AI-assisted offline diagnosis along two complementary paths. Model-side failures include incorrect task decomposition, route selection, argument grounding, state tracking, error recovery, and task closure. Harness-side failures include missing or conflicting Skill guidance, inappropriate tool exposure, retrieval errors, stale Memory, and incomplete feedback formatting. 

The resulting diagnoses enter different update paths. Model-side evidence can guide subsequent task construction, data repair, and optimization objectives. Harness-side evidence can inform candidate revisions to tool-to-Skill mappings, procedural guidance, Memory policies, resource exposure, retrieval mechanisms, and feedback interfaces. Candidate changes remain subject to rule-based checks and version control, while low-confidence, conflicting, or release-critical decisions are routed for human review. No candidate update is applied autonomously to a serving system.

Updates to the model alter the planning behavior observed by the Harness, while updates to the Harness alter the context, resources, and action space encountered by the model. These coupled effects define a governed pathway for model--Harness co-adaptation and, over repeated validated development cycles, toward model--Harness co-evolution. We present this mechanism as a development pathway rather than claiming that sustained autonomous co-evolution has already been established.

\section{Experiments}
\label{sec:experiments}

We evaluate the complete Qwen-Planner-Agent system and its Planner Model from three perspectives: mobile-planning performance and output cost, agentic capabilities beyond mobile tasks, and the contributions of training and Harness support. 
We develop two Planner Models based on Qwen backbones, with model sizes of 35B-A3B and 27B. In our experiments, Qwen-Planner-Model denotes the trained planner without our deployment-time Harness, while Qwen-Planner-Agent denotes the complete system with Harness support.

\subsection{Mobile Planning Performance and Efficiency}
\label{sec:mobile-planning-evaluation}

We assess end-to-end mobile planning on MobilePA-Bench~\citep{zhu2026mobilepa}, comprising \textbf{1,700+ executable tasks}, \textbf{200+ mobile tools}, and \textbf{13 query-task domains}. Tasks start from controlled initial states; completion is verified through tool calls, terminal state changes, or agent behavior. Table~\ref{tab:mobile-cross-model} compares our base models, trained planners, and complete agents with Claude Opus 5, Opus 4.8, and Fable 5~\citep{anthropic2026modelreports}; Gemini 3.6 Flash and 3.1 Pro~\citep{google2026geminimodels}; 
GPT-6 Astra and GPT-5.6 Sol~\citep{openai2026models}; 
Seed 2.1 Pro~\citep{bytedance2026seed21}; Qwen 3.7 Max~\citep{qwen37max} and 3.8 Max~\citep{qwen38}; GLM 5.3, 5.3 Flash, and 5.2~\citep{glm5team2026glm5}; and Kimi K3~\citep{kimi2026k3}. Final evaluation is separate from development evaluations in the AI-for-AI loop (Section~\ref{sec:system-overview}).

\begin{table*}[!t]
\setlength{\tabcolsep}{5pt}
\renewcommand{\arraystretch}{1.25}
\centering
\caption{\textbf{Cross-Model and System Comparison on MobilePA-Bench.} Qwen-Planner-Model reports the trained planner checkpoint without the Harness; Qwen-Planner-Agent includes deployment-time context, Memory, Skill, and feedback support. Bold denotes the highest observed score in each column.}
\label{tab:mobile-cross-model}
\small
\begin{tabular}{l*{6}{c}}
\toprule[1.5pt]
\textbf{Model / System} & \textbf{Access / Size} & \textbf{Overall} & \textbf{Tool Use} & \textbf{Memory} & \textbf{Skills} & \textbf{Sub-agent} \\
\hline
\rowcolor{gray!15}\multicolumn{7}{l}{\textit{\textbf{Closed-Source Models}}} \\ 
GPT 6 Astra & Closed-Source & 76.84 & 75.71 & 74.73 & \textbf{93.25} & 53.93 \\
Claude Opus 5 & Closed-Source & 75.71 & 77.60 & 71.81 & 83.00 & 59.55 \\
Claude Fable 5 & Closed-Source & 74.53 & 77.21 & \textbf{76.33} & 69.00 & \textbf{68.54} \\
Gemini 3.6 Flash & Closed-Source & 69.61 & 74.71 & 69.68 & 65.75 & 51.69 \\
Gemini 3.1 Pro & Closed-Source & 68.16 & 75.38 & 65.69 & 55.75 & 61.80 \\
Claude Opus 4.8 & Closed-Source & 66.05 & 72.40 & 56.38 & 69.25 & 47.19 \\
Seed 2.1 Pro & Closed-Source & 64.89 & 70.29 & 57.18 & 64.00 & 55.06 \\
Qwen 3.7 Max & Closed-Source & 64.89 & 72.31 & 64.10 & 53.75 & 51.69 \\
\hline
\rowcolor{gray!15}\multicolumn{7}{l}
{\textit{\textbf{Open-Source Models}}} \\ 
GLM 5.3 & 744B-A40B & 73.88 & 76.44 & 72.07 & 77.00 & 58.43 \\
Qwen 3.8 Max & 2.4T-A95B & 71.77 & 73.65 & 73.14 & 75.75 & 51.69 \\
Kimi K3 & 2.8T-A104B & 69.64 & 71.54 & 71.01 & 74.75 & 47.19 \\
GLM 5.3 Flash & 320B-A18B & 68.31 & 71.35 & 65.16 & 68.25 & 59.55 \\
GLM 5.2 & 744B-A40B & 66.06 & 73.75 & 60.37 & 58.00 & 55.06 \\
\hline
\rowcolor{gray!15}\multicolumn{7}{l}
{\textit{\textbf{Baselines and Qwen-Planner-Models/Agents}}} \\ 
Qwen Baseline& 35B-A3B & 54.90 & 64.04 & 44.41 & 47.50 & 44.94 \\
\rowcolor{light_bg} Qwen-Planner-Model & 35B-A3B & 64.79 & 66.15 & 61.17 & 71.00 & 52.81 \\
\rowcolor{light_bg} Qwen-Planner-Agent & 35B-A3B & 69.91 & 71.25 & 67.02 & 78.00 & 52.81 \\
Qwen Baseline & 27B & 67.22 & 68.37 & 67.82 & 73.75 & 47.19 \\
\rowcolor{light_bg} Qwen-Planner-Model & 27B & 71.90 & 72.79 & 70.74 & 79.25 & 55.06 \\
\rowcolor{light_bg} Qwen-Planner-Agent & 27B & \textbf{77.05} & \textbf{77.79} & 74.76 & 86.25 & 59.55 \\
\bottomrule[1.5pt]
\end{tabular}
\end{table*}

\paragraph{Performance.}
Qwen-Planner-Agent 27B achieves an Overall score of \textbf{77.05\%} on MobilePA-Bench, ranking first among the evaluated models and agent systems and outperforming GPT 6 Astra (76.84\%) and Claude Opus 5 (75.71\%). Both model sizes improve over their corresponding Qwen baselines: Overall performance rises from 67.22\% to 77.05\% for 27B and from 54.90\% to 69.91\% for 35B-A3B. The contributions of planner training and Harness support are examined separately in the ablation studies. Across individual capabilities, Qwen-Planner-Agent 27B improves over its baseline in Tool Use, Memory, Skills, and Sub-agent coordination, achieving the highest Tool Use score of 77.79\%, with Memory and Skills scores of 74.76\% and 86.25\%, respectively.

\paragraph{Efficiency.}
As shown in Figure~\ref{fig:performance}, Qwen-Planner-Agent combines the highest Overall performance in the cost comparison with the lowest estimated output cost, at \textbf{\$2.41 per 1,000 tasks}, compared with \$3.06--\$67.76 for the other evaluated models. Costs are estimated from mean output tokens per task, including thinking tokens, and the corresponding output-token rates. These results indicate a favorable performance--cost trade-off under the evaluated pricing assumptions. The estimates cover output-token charges only, excluding input tokens, external tool charges, device execution, and additional Harness processing.

\begin{figure}[!t]
    \centering
    \includegraphics[width=0.9\linewidth]{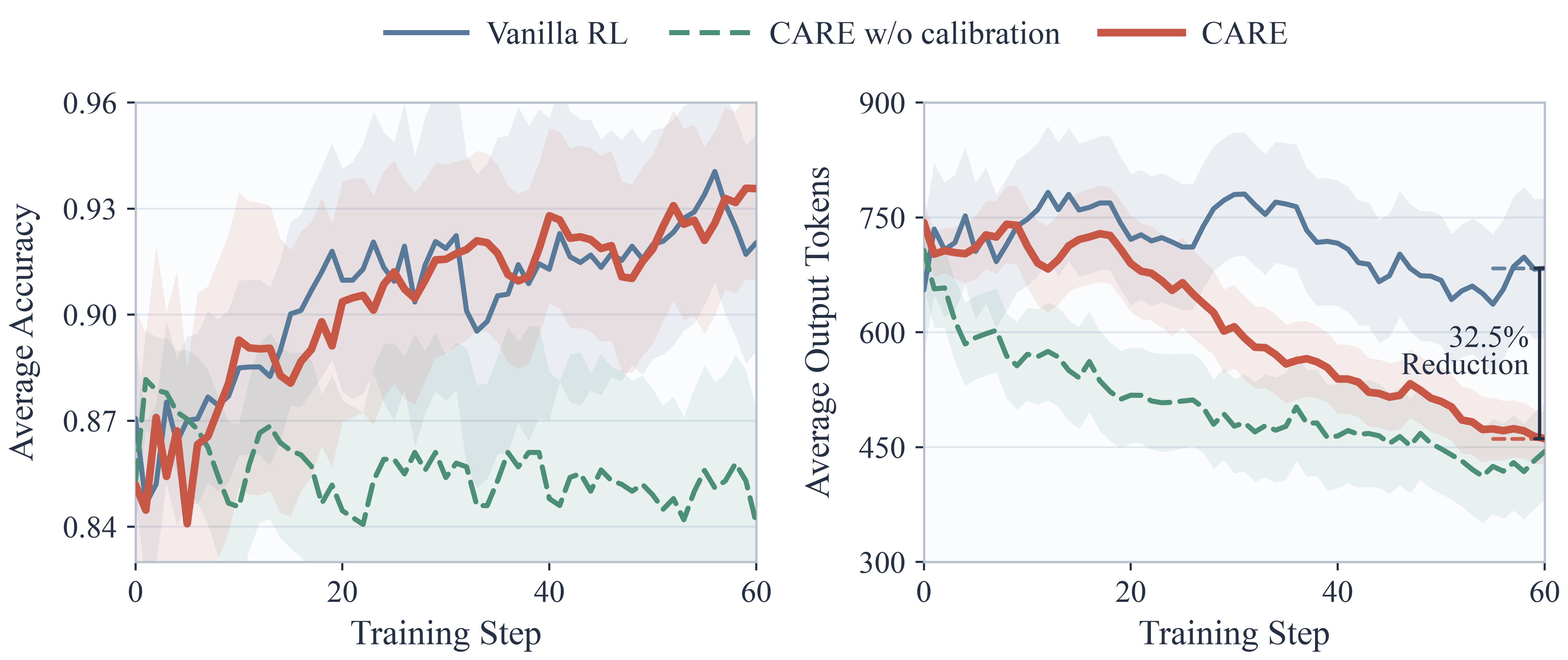}
    \caption{\textbf{Training dynamics of CARE and advantage calibration.} Under the same training setup, full CARE maintains accuracy comparable to Vanilla RL while using 32.5\% fewer output tokens at the final plotted step. CARE without advantage calibration generates even shorter outputs but plateaus at lower accuracy. Lines show smoothed means over four training environments.}
    \label{fig:advantage-calibration-ablation}
\end{figure}

\subsection{Ablation Studies}
\label{sec:ablation-study}

We conduct three complementary analyses to examine training, runtime support, and iterative refinement. First, we evaluate how CARE and advantage calibration affect task success and output efficiency. Second, we measure the contribution of Harness support while holding planner parameters fixed. Finally, we assess whether alternating model training and Harness refinement improves performance on both mobile-planning and general tool-use tasks.

\subsubsection{CARE Training Dynamics and Advantage Calibration}

Using the 27B baseline from the main comparison, we compare Vanilla RL, full CARE, and CARE without quality-preserving advantage calibration under the same training setup across four training environments (Figure~\ref{fig:advantage-calibration-ablation}). The left panel reports average training accuracy, while the right panel reports average output length in tokens. The comparison with Vanilla RL evaluates CARE as a whole, while removing advantage calibration isolates its contribution under the same training setup.

CARE maintains average training accuracy comparable to Vanilla RL while reducing output length by \textbf{32.5\%} at the final shared plotted step, based on the smoothed curves. Without advantage calibration, outputs become shorter but training accuracy remains substantially lower over the plotted interval. This pattern is consistent with the normalization effect analyzed in Section~\ref{sec:rl}: small efficiency differences can acquire disproportionate advantage scale in success-saturated groups, while calibration limits their amplification.

\subsubsection{Harness Contribution under Fixed Checkpoints}
\label{sec:runtime-harness-ablation}

Table~\ref{tab:mobile-cross-model} provides paired evaluations before and after adding the Harness to each trained planner. Holding the 27B checkpoint fixed, the Harness raises Overall from 71.90\% to 77.05\%, with Tool Use increasing from 72.79\% to 77.79\%, Memory from 70.74\% to 74.76\%, Skills from 79.25\% to 86.25\%, and Sub-agent from 55.06\% to 59.55\%. For the 35B-A3B checkpoint, Overall rises from 64.79\% to 69.91\%, while Sub-agent remains unchanged at 52.81\%. These comparisons identify the contribution of deployment-time context and execution support separately from changes in model parameters. The paired evaluations measure the combined effect of Harness support, with Section~\ref{sec:memory-harness-evaluation} further examining its behavior on long-history tasks.

\subsubsection{Toward Model-Harness Co-Evolving}
\label{sec:coevolution-evaluation}

\begin{figure*}[t]
    \centering
    \begin{minipage}{\textwidth}
        \centering
        \captionof{table}{\textbf{Performance Comparison on MobilePA-Internal
        and MCPMark.} All scores are percentages; higher is better.
        Model-harness co-evolution improves the overall scores over
        the model-only baseline by 5.83 and 8.98 percentage points,
        respectively. Co-evolution results are reported after four
        iterations on MobilePA-Internal and three iterations on MCPMark.}
        \label{tab:model-harness-co-evolve}
        \resizebox{0.85\linewidth}{!}{
            \begin{tabular}{lccccc}
                \toprule
                \multirow{2}{*}{\textbf{Method}} & \multicolumn{4}{c}{\textbf{MobilePA-Internal}} & \multirow{2}{*}{\textbf{MCPMark}} \\
                \cmidrule(lr){2-5}
                 & Overall & Tool Use & Long Context
                & Personalized Planning & \\
                \midrule
                Model Only
                & 82.67 & 83.36 & 85.43 & 75.46 & 38.00 \\
                Model+Harness
                & 84.23 & 85.52 & 86.18 & 75.46 & 42.26 \\
                Model--Harness Co-evolution
                & \textbf{88.50} & \textbf{89.76} & \textbf{88.69}
                & \textbf{82.42} & \textbf{46.98} \\
                \bottomrule
            \end{tabular}
        }
    \end{minipage}

    \par\vspace{0.5em}

    \includegraphics[width=\textwidth]
        {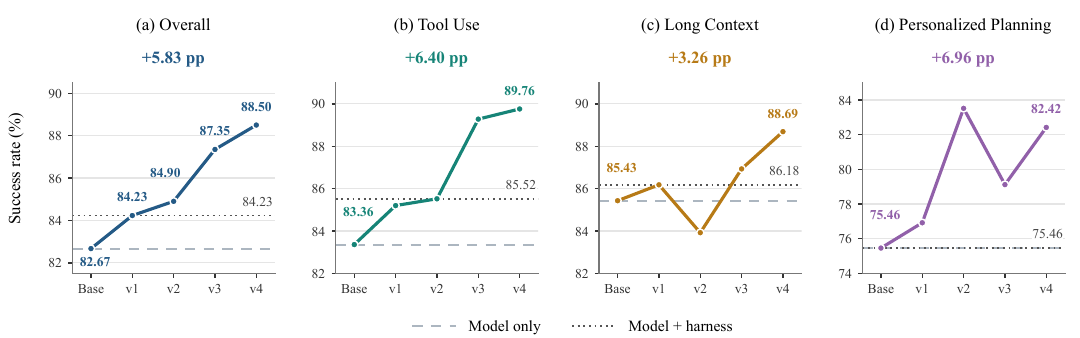}
    \caption{Model--harness co-evolution on MobilePA-Internal across
    four iterations (v1--v4). Base denotes the model-only baseline.
    Horizontal lines mark the two baselines and coincide in (d);
    panel annotations give v4 gains over Base in percentage points.
    Vertical ranges differ across panels.}
    \label{fig:mobilepa-co-evolve}
\end{figure*}

To evaluate the effectiveness of model--harness co-evolution, we use a separate 27B baseline checkpoint from the one used in the main comparison and conduct experiments on MobilePA-Internal and MCPMark.
MobilePA-Internal is a comprehensive internal evaluation set that we construct to cover three task categories: Tool Use, Long Context, and Personalized Planning.
MCPMark~\citep{wu2026mcpmark} is a general-purpose tool-use benchmark spanning five domains, allowing us to examine the framework's applicability beyond mobile-agent tasks.
We compare co-evolution against two baselines: the model alone and the model equipped with harness.
Both experiments organize model training into rounds with a fixed number of training steps. As in Eq.~\ref{eq:coevolution_updates}, each round trains the policy under a fixed harness. The updated policy is then evaluated under the same harness on a held-out evolve set, which serves as the development set for co-evolution rather than an independent final test set.
Feedback from this development set guides harness revision for subsequent training; the evolve tasks and their trajectories are excluded from direct training.

As shown in Table~\ref{tab:model-harness-co-evolve}, Model--Harness Co-evolution improves overall performance from 82.67\% to 88.50\% on MobilePA-Internal after four iterations and from 38.00\% to 46.98\% on MCPMark after three iterations.
These final scores also exceed those of the harness, which achieves 84.23\% on MobilePA-Internal and 42.26\% on MCPMark.
Figure~\ref{fig:mobilepa-co-evolve} shows the iteration-wise progress on MobilePA-Internal: overall performance improves across successive iterations, with Tool Use, Long Context, and Personalized Planning all outperforming both baselines by the final iteration.
These results support the effectiveness of Model--Harness Co-evolution beyond the gains provided by harness alone, across both mobile-agent and general-purpose tool-use tasks.

\subsection{Long-History Memory Evaluation}
\label{sec:memory-harness-evaluation}

\subsubsection{Evaluation Protocol}

We compare without-Harness and with-Harness inference to assess whether persistent memory helps when relevant evidence exceeds the active context. External baselines also include Seed 2.0~\citep{bytedance2026seed20}. Without the Harness, the question and interaction history are placed directly in the model context; models are grouped by a 1M- or 256K-token evaluation budget, and the earliest history is truncated when that budget is exceeded. With the Harness, its Persistent Memory Manager processes the full history and retrieves task-relevant evidence for answering. Each matched Qwen pair uses the same planner checkpoint with different history-access mechanisms; input-token and processing costs vary between the two regimes.

Following the Mem0 evaluation protocol~\citep{mem02026research}, LoCoMo~\citep{maharana2024locomo} and LongMemEval~\citep{wu2024longmemeval} report binary LLM-judge accuracy, while BEAM~\citep{tavakoli2025beam} reports average rubric score. All scores are expressed as percentages, with higher values indicating better performance. Official Mem0 results are external references, not measurements produced with our Harness.

\begin{table*}[!t]
\centering
\caption{\textbf{Long-history Memory Evaluation with and without Harness Support.} Models evaluated without the Harness are grouped by their context budget. Qwen-Planner-Model denotes planner-only inference; Qwen-Planner-Agent denotes the same planner with Harness support. LoCoMo and LongMemEval report binary LLM-judge accuracy, and BEAM reports average rubric score. All scores are reported as percentages (\%). Bold denotes the best observed result in each column; $\dagger$ denotes externally reported Mem0 results, provided for reference rather than evaluated with our Harness. Avg. is the unweighted mean over available benchmark scores; for Mem0, it is computed over the four reported scores.}
\label{tab:memory-benchmarks}
\small
\setlength{\tabcolsep}{5pt}
\renewcommand{\arraystretch}{1.25}
\resizebox{0.95\textwidth}{!}{%
\begin{tabular}{l*{8}{c}}
\toprule[1.5pt]
\textbf{Model / System}
& \textbf{Access / Size}
& \textbf{LoCoMo}
& \makecell{\textbf{LongMem}\\\textbf{Eval}}
& \makecell{\textbf{BEAM}\\\textbf{100K}}
& \makecell{\textbf{BEAM}\\\textbf{500K}}
& \makecell{\textbf{BEAM}\\\textbf{1M}}
& \makecell{\textbf{BEAM}\\\textbf{10M}}
& \textbf{Avg.} \\
\hline
\rowcolor{gray!15}\multicolumn{9}{l}{\textit{\textbf{w/o Memory Harness (1M Context Length Budget)}}} \\
Claude Opus 5 & Closed-Source & \textbf{97.20} & \textbf{97.11} & \textbf{82.19} & \textbf{73.63} & 54.84 & 20.49 & 70.91 \\
GPT-5.6-Sol & Closed-Source & 95.77 & 94.20 & 74.24 & 63.55 & 51.07 & 22.38 & 66.87 \\
Gemini 3.6 Flash & Closed-Source & 95.51 & 94.80 & 70.29 & 65.08 & 45.53 & 18.53 & 64.96 \\
Claude Opus 4.8 & Closed-Source & 95.39 & 94.40 & 68.04 & 59.21 & 41.44 & 16.99 & 62.58 \\
Qwen3.8-Max & 2.4T-A95B & 95.39 & 94.98 & 75.47 & 62.66 & 46.62 & 20.84 & 65.99 \\
GLM-5.2 & 744B-A40B & 95.06 & 93.80 & 71.33 & 64.23 & 48.47 & 25.09 & 66.33 \\
\hline
\rowcolor{gray!15}\multicolumn{9}{l}{\textit{\textbf{w/o Memory Harness (256K Context Length Budget)}}} \\
Kimi K3 & 2.8T-A104B & 95.38 & 95.40 & 77.11 & 44.59 & 40.93 & 21.82 & 62.54 \\
Seed 2.0 & Closed-Source & 94.48 & 88.60 & 63.96 & 37.20 & 37.95 & 24.88 & 57.85 \\
Qwen Baseline & 35B-A3B & 93.38 & 89.40 & 69.31 & 40.92 & 40.64 & 25.64 & 59.88 \\
\rowcolor{light_bg}Qwen-Planner-Model & 35B-A3B & 93.44 & 87.80 & 66.58 & 42.09 & 40.14 & 22.96 & 58.84 \\
Qwen Baseline & 27B & 94.41 & 94.40 & 76.26 & 43.71 & 40.75 & 23.07 & 62.10 \\
\rowcolor{light_bg}Qwen-Planner-Model & 27B & 94.35 & 95.00 & 72.32 & 42.01 & 40.12 & 21.99 & 60.97 \\
\hline
\rowcolor{gray!15}\multicolumn{9}{l}{\textit{\textbf{w/ Memory Harness}}} \\
Mem0 (Official)$^\dagger$ & Closed-Source & 92.50 & 94.40 & -- & -- & 64.10 & 48.60 & 74.90 \\
Qwen Baseline & 35B-A3B & 93.18 & 90.00 & 64.93 & 63.87 & 65.17 & 56.68 & 72.31 \\
\rowcolor{light_bg}Qwen-Planner-Agent & 35B-A3B & 93.44 & 86.55 & 69.93 & 65.65 & 68.21 & 57.23 & 73.50 \\
Qwen Baseline & 27B & 94.93 & 95.80 & 78.00 & 71.42 & \textbf{73.97} & 65.50 & 79.94 \\
\rowcolor{light_bg}Qwen-Planner-Agent & 27B & 94.93 & 96.60 & 75.35 & 72.20 & 73.71 & \textbf{67.24} & \textbf{80.01} \\
\bottomrule[1.5pt]
\end{tabular}%
}
\end{table*}

\subsubsection{Results Across History Lengths}

\paragraph{Direct-context performance as history grows.}
Direct-context inference remains strong on LoCoMo, LongMemEval, and BEAM-100K, but performance declines as history grows. Claude Opus 5 provides the strongest direct-context result through BEAM-1M, reaching 73.63\% on BEAM-500K and 54.84\% on BEAM-1M. Under the truncation protocol described above, no direct-context model exceeds 25.64 on BEAM-10M. Qwen-Planner-Model 27B remains competitive in the direct 256K setting, scoring 94.35\% on LoCoMo and 95.00\% on LongMemEval.

\paragraph{Matched Harness gains across history lengths.}
The benefit of explicit memory becomes pronounced on the longer BEAM settings. Across all four matched Qwen model pairs reported in Table~\ref{tab:memory-benchmarks}, the Harness improves every result on BEAM-500K, BEAM-1M, and BEAM-10M, raising mean scores from 42.18\% to 68.29\%, from 40.41\% to 70.27\%, and from 23.42\% to 61.66\%, respectively. By contrast, the mean scores on LoCoMo, LongMemEval, and BEAM-100K change only from 93.90\% to 94.12\%, from 91.65\% to 92.24\%, and from 71.12\% to 72.05\%, respectively, and several individual pairs show small regressions. For Qwen-Planner-Model 27B, adding the Harness raises BEAM-500K from 42.01\% to 72.20\%, BEAM-1M from 40.12\% to 73.71\%, and BEAM-10M from 21.99\% to 67.24\%.

\paragraph{Comparison, trade-offs, and interpretation.}
The strongest with-Harness results are split across the longest settings: Qwen Baseline (27B) with the Harness achieves the best BEAM-1M score at 73.97\%, while Qwen-Planner-Agent 27B achieves the best BEAM-10M score at 67.24\% and scores 73.71\% on BEAM-1M. For reference, the best direct-context results on BEAM-1M and BEAM-10M are 54.84\% and 25.64\%, and the officially reported Mem0 results are 64.10\% and 48.60\%. The Mem0 scores are external reference results. Across the matched Qwen comparisons, Harness gains are concentrated in the longer-history settings, while changes on shorter-history benchmarks vary across model pairs. Section~\ref{sec:behavioral-analysis} illustrates how evidence retrieval and reconciliation support long-history reasoning.

\subsection{General Agentic Capability Evaluation}
\label{sec:generalization-evaluation}

Qwen-Planner-Model improves agentic performance beyond mobile planning while largely preserving general capabilities. Table~\ref{tab:general-agent-evaluation} compares planner-only inference with the corresponding Qwen baseline on Claw-Eval~\citep{ye2026claw}, BFCL-v4~\citep{patil2025bfcl}, MCP-Atlas~\citep{bandi2026mcpatlas}, Tau-3~\citep{sierra2026tau3}, Toolathlon~\citep{li2025toolathlon}, SWE-Multilingual~\citep{khandpur2025swemultilingual}, and SWE-Pro~\citep{deng2025swepro}. The 35B-A3B and 27B variants improve on six and five of these seven agentic benchmarks, respectively, demonstrating stronger capabilities across function calling, multi-tool interaction, and coding tasks, although gains are not uniform across benchmarks. Meanwhile, performance on MMLU-Redux~\citep{gema2024mmluredux}, C-Eval~\citep{huang2023ceval}, and IFEval~\citep{zhou2023ifeval} remains close to the baselines, with small decreases. Together, these results show that our training recipe strengthens capabilities applicable to non-mobile agentic tasks without substantially compromising general reasoning and instruction following.

\begin{table*}[!t]
\centering
\caption{\textbf{Performance Beyond Mobile Planning.} Benchmarks are grouped into agentic capability and general capability, with coding-agent evaluations included in the former. Each Qwen-Planner-Model checkpoint is compared with its corresponding base model. Avg. denotes the arithmetic average of the seven agentic or three general benchmarks within each group. All scores are percentages (\%).}
\label{tab:general-agent-evaluation}
\scriptsize
\setlength{\tabcolsep}{2.8pt}
\renewcommand{\arraystretch}{1.4}
\resizebox{0.99\textwidth}{!}{%
\begin{tabular}{l*{9}{c}@{\hspace{5pt}}*{4}{c}}
\toprule[1.5pt]
\multirow{3}{*}{\textbf{Model}}
& \multirow{3}{*}{\textbf{Size}}
& \multicolumn{8}{c}{{\textbf{Agentic Capability}}}
& \multicolumn{4}{c}{{\textbf{General Capability}}} \\
\cmidrule(lr){3-10}\cmidrule(lr){11-14}
& &  \makecell{Claw-\\Eval}
& \makecell{BFCL-\\v4}
& \makecell{MCP-\\Atlas}
& Tau-3
& \makecell{Tool-\\athlon}
& \makecell{SWE-\\Multil.}
& \makecell{SWE-\\Pro}
& \cellcolor{gray!15}\textbf{Avg.}
& \makecell{MMLU-\\Redux}
& C-Eval
& IFEval
& \cellcolor{gray!15}\textbf{Avg.} \\
\midrule
Qwen Baseline & 35B-A3B
& 72.83 & 60.63 & 59.40 & 67.94 & 29.30 & 65.67 & 48.43
& \cellcolor{gray!15}57.74
& 89.67 & 90.45 & 90.94 
& \cellcolor{gray!15}90.35 \\
\rowcolor{light_bg}
Qwen-Planner-Model & 35B-A3B
& \textbf{74.51} & \textbf{66.99} & \textbf{68.69} & 67.06 & \textbf{37.15} & \textbf{71.67} & \textbf{50.07}
& \cellcolor{gray!15}\textbf{62.31}
& 89.53 & 90.39 & 89.83 
& \cellcolor{gray!15}89.92 \\
\hline
Qwen Baseline & 27B
& 82.48 & 67.06 & 71.05 & 74.27 & 39.70 & 76.65 & 61.70
& \cellcolor{gray!15}67.91
& 88.73 & 89.74 & 91.68 
& \cellcolor{gray!15}90.05 \\
\rowcolor{light_bg}
Qwen-Planner-Model & 27B
& 79.58 & \textbf{73.60} & \textbf{73.25} & \textbf{74.95} & \textbf{44.02} & 74.89 & \textbf{64.71}
& \cellcolor{gray!15}\textbf{69.29}
& 88.67 & 89.30 & 91.13 
& \cellcolor{gray!15}89.70 \\
\bottomrule[1.5pt]
\end{tabular}}

\end{table*}

\subsection{Qualitative Analysis}
\label{sec:behavioral-analysis}

To complement the quantitative evaluation in Section~\ref{sec:mobile-planning-evaluation}, we examine six execution traces that illustrate how Qwen-Planner-Agent coordinates planning and execution. In the unified runtime described in Section~\ref{sec:model-harness-runtime}, the Harness supplies context and execution feedback, while the Planner Model selects and revises actions. The cases progress from reconciling persistent Memory and interaction history with current state to enforcing Skill-guided prerequisites, selecting conditional actions, and recovering from tool failures. Cross-application sub-agent coordination extends these dependencies across application boundaries. Together, the trajectories show how Qwen-Planner-Agent preserves task constraints while adapting execution to observed results.

\begin{figure}[!htbp]
    \centering
    \captionsetup{font=small,labelfont=bf,skip=5pt}
    \includegraphics[width=\linewidth,height=0.9\textheight,keepaspectratio]{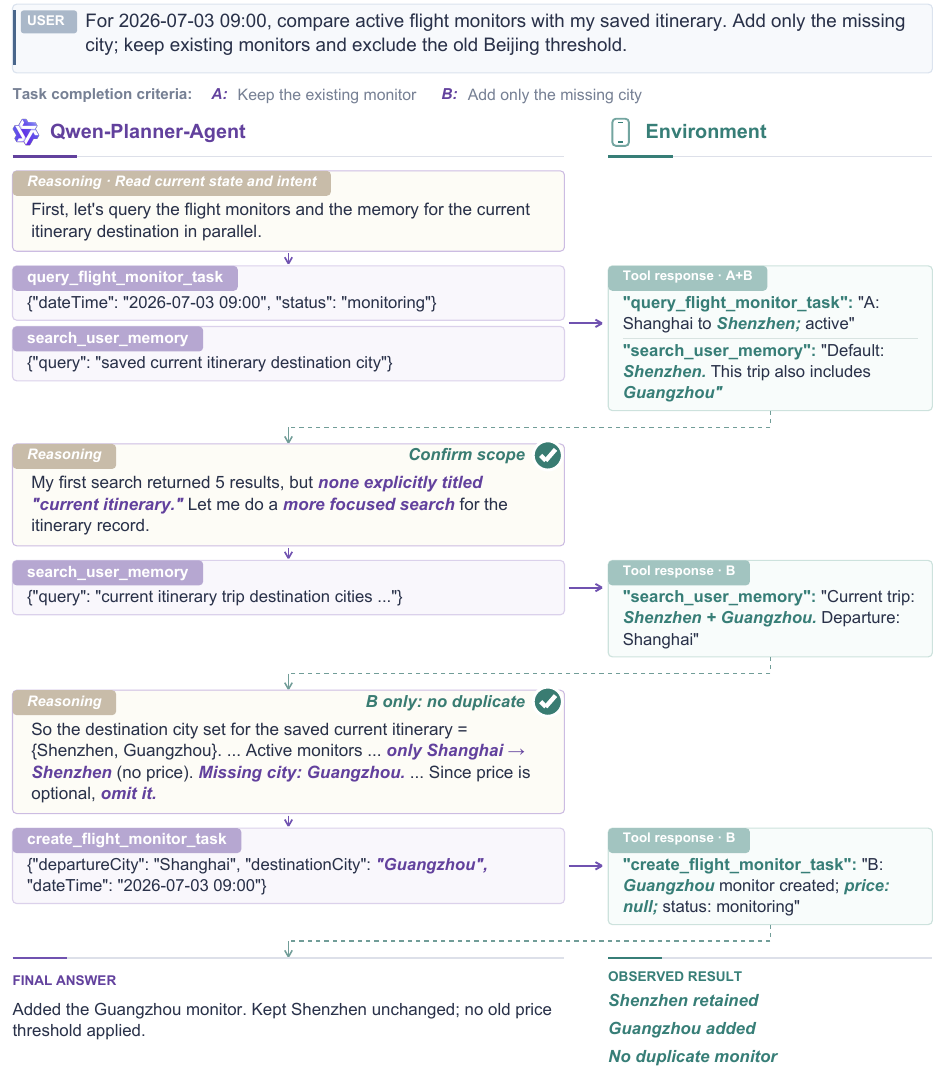}
    \caption{\textbf{Reconciling persistent memory with active task state.} The saved itinerary includes Shenzhen and Guangzhou, while an active monitor already covers Shenzhen (A). Qwen-Planner-Agent adds only Guangzhou (B), preserving the existing monitor and excluding the old Beijing price threshold.}
    \label{fig:memory_state_reconciliation}
\end{figure}

\Needspace{6\baselineskip}
\paragraph{Reconciling persistent memory with live state.} Personalized execution requires reconciling remembered intent with the current application state. In Figure~\ref{fig:memory_state_reconciliation}, Qwen-Planner-Agent retrieves the active flight monitors and saved itinerary, then confirms that the trip covers both Shenzhen and Guangzhou. Because Shenzhen is already monitored, it creates only the missing Guangzhou monitor, leaving the existing record and its settings intact. The case illustrates how memory and live state jointly determine the required action while preventing duplicate work.

\begin{figure}[!htbp]
    \centering
    \captionsetup{font=small,labelfont=bf,skip=5pt}
    \includegraphics[width=\linewidth,height=0.9\textheight,keepaspectratio]{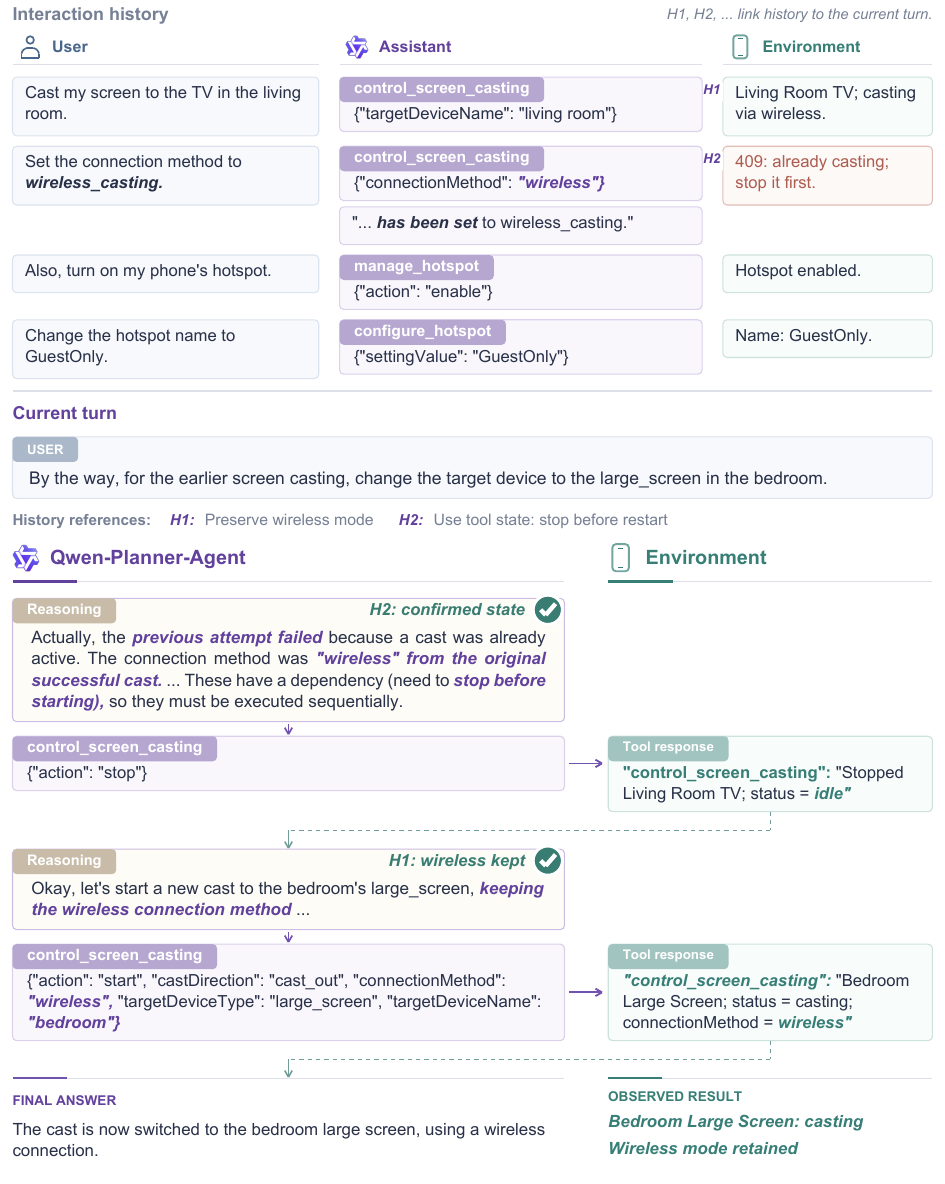}
    \caption{\textbf{State-grounded tool use across dialogue turns.} H1 records the established wireless mode, while H2 records that a cast is still active. Qwen-Planner-Agent follows these states, waits for the active session to stop, then starts the bedroom cast. The final return confirms both the revised destination and the retained connection method.}
    \label{fig:vis_multiturn_casting_revision}
\end{figure}

\Needspace{6\baselineskip}
\paragraph{Resuming tasks from verified execution state.} After two hotspot requests, Qwen-Planner-Agent resumes casting from tool-confirmed state despite a misleading success acknowledgement (Figure~\ref{fig:vis_multiturn_casting_revision}). The history records an active wireless session and a rejected second start (H1--H2), establishing that the existing session must first be stopped. It stops the active session, waits for idle, and starts the bedroom cast while preserving wireless mode.

\begin{figure}[!htbp]
    \centering
    \captionsetup{font=small,labelfont=bf,skip=5pt}
    \includegraphics[width=\linewidth,height=0.9\textheight,keepaspectratio]{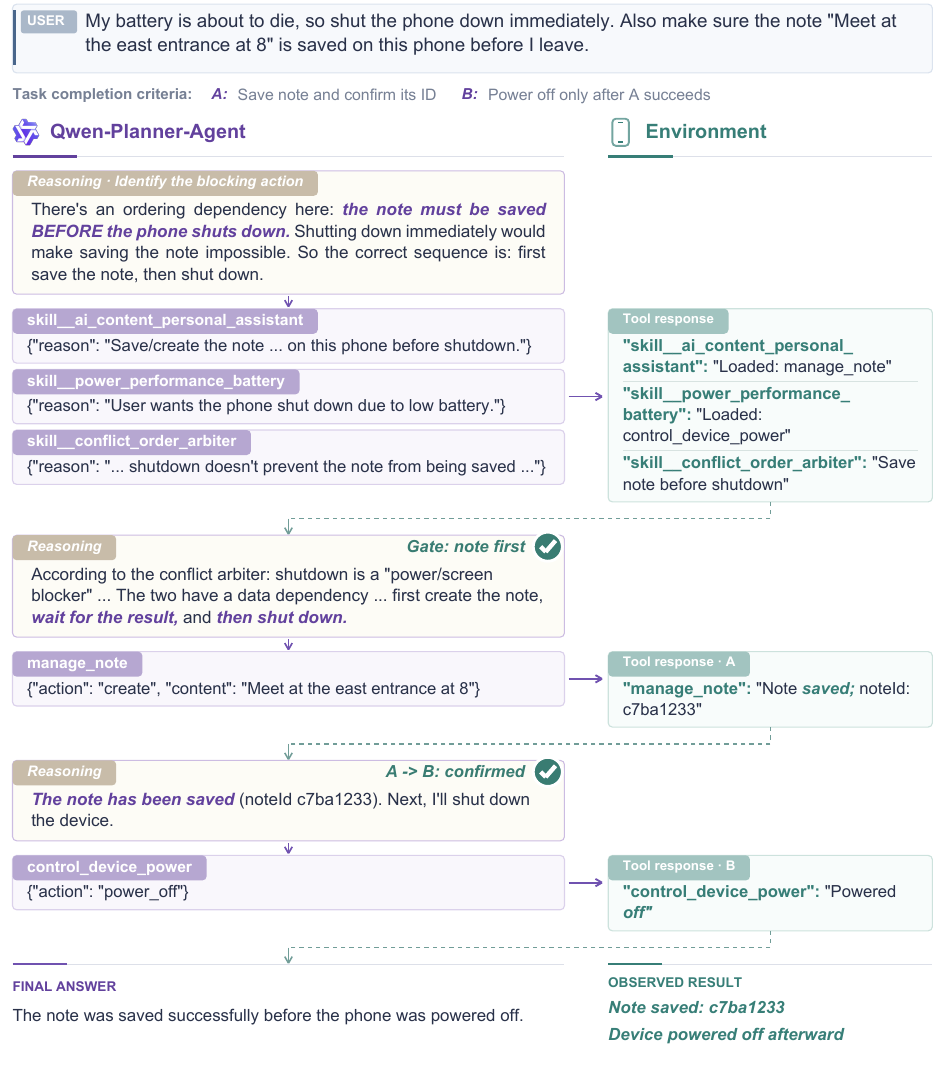}
    \caption{\textbf{Skill-guided prerequisite preservation before shutdown.} Qwen-Planner-Agent first obtains a persistent identifier for the saved note (A) and only then issues power-off (B). The returned states confirm that the retrieved procedural constraint is realized in execution and that both parts of the request complete.}
    \label{fig:vis_mobile_skill_2}
\end{figure}

\Needspace{6\baselineskip}
\paragraph{Enforcing execution prerequisites with Skills.} An urgent shutdown request also requires a note to be saved, creating a dependency that must be resolved before the device becomes unavailable. In Figure~\ref{fig:vis_mobile_skill_2}, Qwen-Planner-Agent loads the note, power, and conflict-arbitration Skills and waits for the saved note's identifier before issuing power-off. Retrieved guidance thus becomes an execution prerequisite that preserves task completion before a blocking action.

\begin{figure}[!htbp]
    \centering
    \captionsetup{font=small,labelfont=bf,skip=5pt}
    \includegraphics[width=\linewidth,height=0.9\textheight,keepaspectratio]{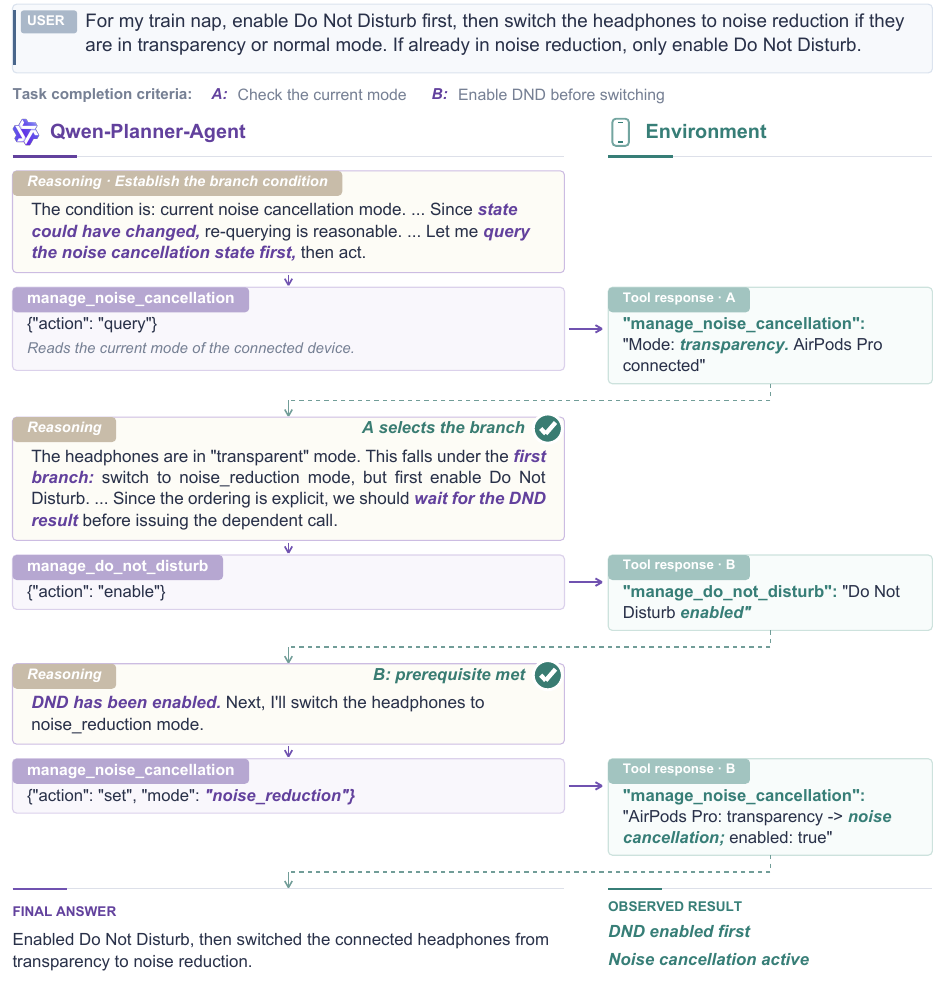}
    \caption{\textbf{Conditional tool use under an ordering constraint.} The observed transparency mode selects the branch that requires a headphone adjustment (A). Qwen-Planner-Agent first enables Do Not Disturb, then uses its successful return to proceed with noise reduction (B). Both requested settings are confirmed by the tool results.}
    \label{fig:tool_conditional_planning}
\end{figure}

\Needspace{6\baselineskip}
\paragraph{Conditioning tool execution on observed state.} Conditional requests require the planner to resolve both whether an action is needed and when it may be issued. In Figure~\ref{fig:tool_conditional_planning}, Qwen-Planner-Agent queries the connected headphones and selects the branch corresponding to transparency mode. It then enables Do Not Disturb and waits for the successful return before switching to noise reduction. The trajectory combines state-dependent branching with an explicit ordering constraint, translating the user's conditional request into a sequence of native tool calls.

\begin{figure}[!htbp]
    \centering
    \captionsetup{font=small,labelfont=bf,skip=5pt}
    \includegraphics[width=\linewidth,height=0.80\textheight,keepaspectratio]{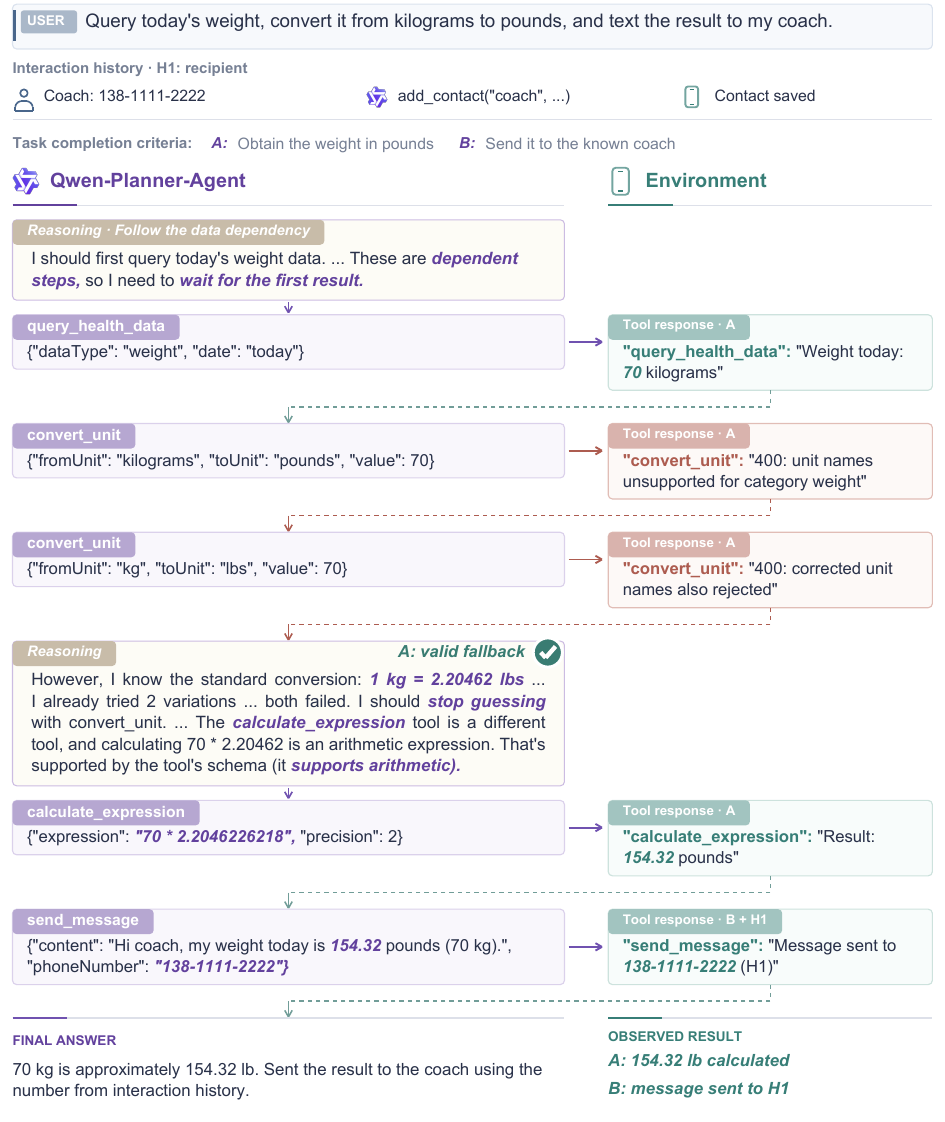}
    \caption{\textbf{Tool substitution for workflow recovery.} After two conversion attempts fail, Qwen-Planner-Agent uses the calculator to obtain 154.32 pounds (A) and sends the result to the coach (B). H1 links the earlier saved contact to the final message. All five current tool-call rounds are shown; the successful calculation and message return establish completion of the recovered workflow.}
    \label{fig:tool_failure_recovery}
\end{figure}

\Needspace{6\baselineskip}
\paragraph{Recovering execution through tool substitution.} An alternative tool can sustain execution when the original operation fails. In Figure~\ref{fig:tool_failure_recovery}, the unit-conversion tool rejects both full unit names and their abbreviations. Qwen-Planner-Agent uses the standard conversion factor to compute 154.32 pounds from the observed 70 kilograms, then sends the result to the coach identified in the interaction history. This substitution preserves the plan's data dependency, allowing the remaining communication step to complete.

\begin{figure}[!htbp]
    \centering
    \captionsetup{font=small,labelfont=bf,skip=5pt}
    \includegraphics[width=\linewidth,height=0.84\textheight,keepaspectratio]{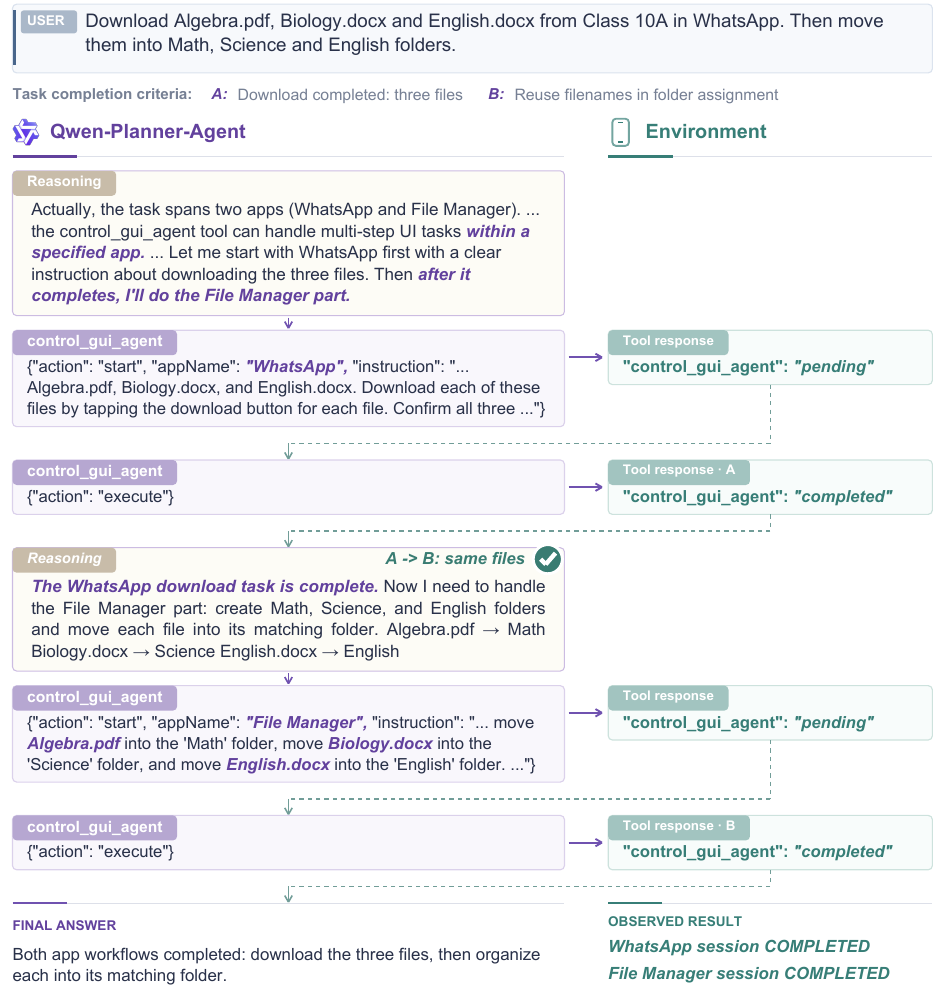}
    \caption{\textbf{Dependency-preserving sub-agent coordination.} Qwen-Planner-Agent waits for the WhatsApp download result (A), then passes the same three filenames and their destination folders to the File Manager session (B). The two session returns and the task-state check confirm completion across the application boundary.}
    \label{fig:vis_mobile_sub_agent}
\end{figure}

\Needspace{6\baselineskip}
\paragraph{Preserving dependencies across sub-agent handoffs.} Sub-agent orchestration requires handoffs to be conditioned on execution results and to preserve the artifacts needed by the next stage. In Figure~\ref{fig:vis_mobile_sub_agent}, Qwen-Planner-Agent waits for the WhatsApp session to report three completed downloads before launching the File Manager session. It then transfers the returned filenames together with their subject-folder assignments into the second instruction. The A-to-B transition is therefore grounded in a completed result rather than the order of two prompts. Both session returns and the recorded task state confirm that the cross-application workflow reaches the requested organization.

\section{Related Work}
\label{sec:related-work}

\subsection{AI-Assisted AI Development}

AI-assisted data production provides one route for improving subsequent models. Self-Instruct generates instruction-tuning examples, while Cosmopedia extends synthetic generation to pretraining corpora~\citep{wang2022selfinstruct,gunther2024cosmopedia}. Feedback can make this process more targeted: LLM2LLM augments examples that a student model fails on, and AutoIF uses code-based verification and execution feedback to filter instruction-following data~\citep{lee2024llm2llm,dong2024autoif}. Our data pipeline applies feedback-guided refinement to interactive planning. AI-assisted task construction and failure diagnosis guide new tasks and sampling adjustments, while automated collection and curation produce executable training experience.

AI also participates in optimizing the artifacts and workflows used to develop models. OPRO, TextGrad, and DSPy use evaluation signals to improve candidate solutions, textual components, or language-model programs~\citep{yang2023opro,yuksekgonul2024textgrad,khattab2023dspy}. Reward-generation methods such as Eureka and CARD extend this assistance to reinforcement learning~\citep{ma2023eureka,sun2024card}, while AutoML-Agent and the AI Scientist address broader development and research workflows~\citep{trirat2024automlagent,lu2024aiscientist}. These directions motivate our systems-level investigation: how can execution experience guide connected improvements in data, training, and deployment? We study this question through the development of Qwen-Planner-Agent, using diagnosed failures to inform training tasks, learning configurations, and runtime revisions within one iterative lifecycle.

\subsection{Mobile Agents and Interactive Environments}

Mobile-agent systems connect high-level requests to application operations through different interfaces. AppAgent learns application-specific interaction knowledge from exploration or demonstrations, while Mobile-Agent-E separates planning from visual execution and retains reusable experience~\citep{zhang2025appagent,wang2025mobile}. AppWorld instead exposes structured application APIs and evaluates tasks through environment state~\citep{trivedi2024appworld}. GUI interaction and structured tools offer complementary access to applications; neither interface alone determines whether an agent can track dependencies, maintain context, recover from errors, and verify a complex task's completion.

Qwen-Planner-Agent focuses on this planning problem rather than introducing a new screen-grounding architecture. MobilePA-Bench measures relevant capabilities through stateful execution across Tool Use, Memory, Skills, and Sub-agent~\citep{zhu2026mobilepa}. To develop these capabilities at scale, we combine reproducible programmatic sandboxes, LLM-simulated long-tail interactions, and selected real-device sessions. The emphasis is on using their complementary feedback to support data collection and online learning, while the Planner Model and Harness coordinate execution under changing tasks and resources.

\subsection{Agentic Reinforcement Learning}

Verifiable-reward optimization provides a foundation for training interactive planners. GRPO normalizes rewards within sampled groups without a learned critic, and DAPO and GSPO further develop group-based policy optimization~\citep{shao2024deepseekmath,yu2026dapo,gspo}. In interactive tasks, actions also change the environment and determine subsequent observations, complicating exploration and credit assignment~\citep{zhang2025agenticrl}. Process supervision, partial rollouts, and branching exploration address aspects of these challenges by providing intermediate learning signals or reusing interaction experience~\citep{lightman2024let,kimi2025k1.5,dong2025arpo}.

Mobile RL additionally faces costly resets and variable execution conditions. DigiRL combines offline initialization with online Android interaction; MobileRL uses difficulty-adaptive replay and curriculum filtering; Mobile-R1 trains through multi-turn interaction; and PhoneBuddy combines resettable mock applications with real-app training~\citep{bai2024digirl,xu2026mobilerl,gu2025mobile,tang2026phonebuddy}. Our approach combines hybrid-environment online agentic RL with competence-adaptive optimization. CARE changes the reward emphasis as task success improves and calibrates advantages to prevent small efficiency differences from dominating learning. A bounded LLM-based controller configures predefined scheduling parameters using training and development-set feedback. This separates feedback-driven configuration from the rule-based reward and advantage computations applied during training.

\subsection{Agent Harnesses and Model--Harness Co-evolution}

Memory and skills provide complementary runtime resources for long-horizon agents. External-memory methods retain episodes, reflections, managed context, or multimodal evidence for later retrieval~\citep{shinn2023reflexion,packer2023memgpt,xu2026mem,chen2025telemem,du2026mom}. Skill-oriented methods retain executable procedures, induced workflows, and reusable task knowledge~\citep{wang2023voyager,wang2024agent,lee2023explore,wang2025mobile,chen2026skilldroid}. These resources motivate a Harness that selects relevant historical evidence and procedural guidance for the current task, rather than requiring the Planner Model to encode every changing fact or operating rule in its parameters.

Harness optimization makes this runtime support itself an object of improvement. AutoHarness synthesizes code harnesses from environment feedback, Meta-Harness searches over Harness code using prior scores and traces, Natural-Language Agent Harnesses expresses control logic as editable instructions, and MemoHarness uses prior execution experience to adapt Harness control dimensions~\citep{lou2026autoharness,lee2026meta,pan2026natural,huang2026memoharness}. Our focus is the interaction between Harness revision and model learning: a revised Harness changes the context and experience used in subsequent policy training, while the updated model's behavior informs the next revision. We implement this coupling through alternating reinforcement learning and LLM-based instruction editing, using held-out development feedback. The multi-round experiments examine this process in mobile-planning and general tool-use settings, providing evidence for iterative model--Harness improvement within the evaluated scope.

\section{Conclusion}
\label{sec:conclusion}

In this report, we present Qwen-Planner-Agent, a unified Planner Model--Harness agentic system developed through a closed-loop feedback-driven AI-for-AI framework. Execution experience guides AI-assisted improvements in data production, training, and deployment. Qwen-Planner-Agent achieves the highest Overall score on MobilePA-Bench among the evaluated systems at a low estimated output cost, while Qwen-Planner-Model improves on most matched non-mobile agentic benchmarks and largely preserves general capabilities. Multi-round experiments further support alternating model training and Harness refinement within the evaluated settings. Together, these results provide a concrete practice of using AI to help develop more capable agents.
Future work will focus on reducing manual intervention in data refinement, training adaptation, and Harness updates, enabling more scalable and efficient iterations of agent capabilities. We aim to automate more of this development lifecycle while retaining human oversight at safety-critical decision points.

\clearpage
\section{Contributors}

\textcolor{stargold}{$\bigstar\,\bigstar\,\bigstar\,\bigstar\,\bigstar$}

Tingyu Qu$^*$, Weigao Sun$^*$, Yuecheng Liu$^*$, Yucheng Zhao$^*$, Yi Zhu$^*$, Yifeng Ding, Qiyi Wang

\textcolor{stargold}{$\bigstar\,\bigstar\,\bigstar\,\bigstar$}

Sihan Cao, Pengkun Jiao, Hanlei Xie,  Xiongwei Wu

\textcolor{stargold}{$\bigstar\,\bigstar\,\bigstar$}

Qichao Wang, Haodong Zhang, Jiajun Liu, Yuhao Wang, Yuqing Xie, Junpeng Zhao, Long Chen, Ming Ma, Sihan Yang, Ziwang Zhao, Yanhao Jia, Liangquan Gong

\textbf{Supervisory panel} 

Feida Zhu, Yiran Zhong, Steven Hoi

\textbf{Collaborator}

Cheng Shi, Chengqi Wu, Chuan Yi, Feng Wang, Fei Yuan, Ju Huang, Jiamang Wang, Sheng Guo, Shaopan Xiong, Siran Yang, Yan Chen,  Zihan Liu

\renewcommand{\thefootnote}{*}
\footnotetext[0]{Equal contribution. Alphabetic order.}

{
    \clearpage
    \bibliographystyle{unsrt}
    \bibliography{references} 

@article{bai2024digirl,
  title={Digirl: Training in-the-wild device-control agents with autonomous reinforcement learning},
  author={Bai, Hao and Zhou, Yifei and Cemri, Mert and Pan, Jiayi and Suhr, Alane and Levine, Sergey and Kumar, Aviral},
  journal={Advances in Neural Information Processing Systems},
  volume={37},
  pages={12461--12495},
  year={2024}
}

@inproceedings{dong2025arpo,
  title={Agentic reinforced policy optimization},
  author={Dong, Guanting and Mao, Hangyu and Ma, Kai and Bao, Licheng and Chen, Yifei and Wang, Zhongyuan and Chen, Zhongxia and Du, Jiazhen and Wang, Huiyang and Zhang, Fuzheng and others},
  booktitle={International Conference on Learning Representations},
  volume={2026},
  pages={16981--17017},
  year={2026}
}

@article{zhu2026mobilepa,
  title={MobilePA-Bench: Benchmarking Mobile Planner Agents on Complex Real-World Tasks},
  author={Zhu, Yi and Wu, Xiongwei and Wang, Qiyi and Qu, Tingyu and Liu, Jiajun and Cao, Sihan and Chen, Long and Sun, Weigao and Zhu, Feida and Zhong, Yiran and others},
  journal={arXiv preprint arXiv:2608.23035},
  year={2026}
}

@article{shao2024deepseekmath,
  title={Deepseekmath: Pushing the limits of mathematical reasoning in open language models},
  author={Shao, Zhihong and Wang, Peiyi and Zhu, Qihao and Xu, Runxin and Song, Junxiao and Bi, Xiao and Zhang, Haowei and Zhang, Mingchuan and Li, YK and Wu, Yang and others},
  journal={arXiv preprint arXiv:2402.03300},
  year={2024}
}

@misc{gspo,
  title         = {Group Sequence Policy Optimization},
  author        = {Zheng, Chujie and Liu, Shixuan and Li, Mingze and Chen, Xiong-Hui and Yu, Bowen and Gao, Chang and Dang, Kai and Liu, Yuqiong and Men, Rui and Yang, An and Zhou, Jingren and Lin, Junyang},
  year          = {2025},
  eprint        = {2507.18071},
  archivePrefix = {arXiv},
  primaryClass  = {cs.LG},
  doi           = {10.48550/arXiv.2507.18071},
  url           = {https://arxiv.org/abs/2507.18071}
}

@article{gdpo,
  title={Gdpo: Group reward-decoupled normalization policy optimization for multi-reward rl optimization},
  author={Liu, Shih-Yang and Dong, Xin and Lu, Ximing and Diao, Shizhe and Belcak, Peter and Liu, Mingjie and Chen, Min-Hung and Yin, Hongxu and Wang, Yu-Chiang Frank and Cheng, Kwang-Ting and others},
  journal={arXiv preprint arXiv:2601.05242},
  year={2026}
}

@article{yu2026dapo,
  title={Dapo: An open-source llm reinforcement learning system at scale},
  author={Yu, Qiying and Zhang, Zheng and Zhu, Ruofei and Yuan, Yufeng and Zuo, Xiaochen and Yue, Yu and Dai, Weinan and Fan, Tiantian and Liu, Gaohong and Liu, Lingjun and others},
  journal={Advances in Neural Information Processing Systems},
  volume={38},
  pages={113222--113244},
  year={2026}
}

@inproceedings{lightman2024let,
  title={Let's verify step by step},
  author={Lightman, Hunter and Kosaraju, Vineet and Burda, Yuri and Edwards, Harrison and Baker, Bowen and Lee, Teddy and Leike, Jan and Schulman, John and Sutskever, Ilya and Cobbe, Karl},
  booktitle={International Conference on Learning Representations},
  volume={2024},
  pages={39578--39601},
  year={2024}
}

@inproceedings{zhang2025appagent,
  title={Appagent: Multimodal agents as smartphone users},
  author={Zhang, Chi and Yang, Zhao and Liu, Jiaxuan and Li, Yanda and Han, Yucheng and Chen, Xin and Huang, Zebiao and Fu, Bin and Yu, Gang},
  booktitle={Proceedings of the 2025 CHI conference on human factors in computing systems},
  pages={1--20},
  year={2025}
}

@inproceedings{trivedi2024appworld,
  title={Appworld: A controllable world of apps and people for benchmarking interactive coding agents},
  author={Trivedi, Harsh and Khot, Tushar and Hartmann, Mareike and Manku, Ruskin and Dong, Vinty and Li, Edward and Gupta, Shashank and Sabharwal, Ashish and Balasubramanian, Niranjan},
  booktitle={Proceedings of the 62nd Annual Meeting of the Association for Computational Linguistics (Volume 1: Long Papers)},
  pages={16022--16076},
  year={2024}
}

@inproceedings{xu2026mobilerl,
  title={Mobilerl: Online agentic reinforcement learning for mobile gui agents},
  author={Xu, Yifan and Liu, Xiao and Liu, Xinghan and Fu, Jiaqi and Huang, Jiayu and Zhang, Hanchen and Jing, Bohao and Zhang, Shudan and Wang, Yuting and Dong, Yuxiao and others},
  booktitle={International Conference on Learning Representations},
  volume={2026},
  pages={35282--35315},
  year={2026}
}

@article{kimi2025k1.5,
  title={Kimi k1. 5: Scaling reinforcement learning with llms},
  author={Team, Kimi and Du, Angang and Gao, Bofei and Xing, Bowei and Jiang, Changjiu and Chen, Cheng and Li, Cheng and Xiao, Chenjun and Du, Chenzhuang and Liao, Chonghua and others},
  journal={arXiv preprint arXiv:2501.12599},
  year={2025}
}

@article{gu2025mobile,
  title={Mobile-r1: Towards interactive reinforcement learning for vlm-based mobile agent via task-level rewards},
  author={Gu, Jihao and Ai, Qihang and Wang, Yingyao and Bu, Pi and Xing, Jingxuan and Zhu, Zekun and Jiang, Wei and Wang, Ziming and Zhao, Yingxiu and Zhang, Ming-Liang and others},
  journal={arXiv e-prints},
  pages={arXiv--2506},
  year={2025}
}

@article{tang2026phonebuddy,
  title={PhoneBuddy: Training Open Models for Agentic Phone Use},
  author={Tang, Zhengyang and Lai, Xin and Lyu, Pengyuan and Wang, Xinyuan and Bai, Tianyi and Li, Chenxin and Guo, Yiduo and Shen, Huawen and Liu, Yuxuan and Li, Junyi and others},
  journal={arXiv e-prints},
  pages={arXiv--2606},
  year={2026}
}

@article{shinn2023reflexion,
  title={Reflexion: Language agents with verbal reinforcement learning},
  author={Shinn, Noah and Cassano, Federico and Gopinath, Ashwin and Narasimhan, Karthik and Yao, Shunyu},
  journal={Advances in neural information processing systems},
  volume={36},
  pages={8634--8652},
  year={2023}
}

@article{packer2023memgpt,
  title={Memgpt: Towards llms as operating systems},
  author={Packer, Charles and Wooders, Sarah and Lin, Kevin and Fang, Vivian and Patil, Shishir G and Stoica, Ion and Gonzalez, Joseph E},
  journal={arXiv preprint arXiv:2310.08560},
  year={2023}
}

@inproceedings{du2026mom,
  title={Mom: Linear sequence modeling with mixture-of-memories},
  author={Du, Jusen and Sun, Weigao and Lan, Disen and Hu, Jiaxi and Tao, Zhang and Cheng, Yu},
  booktitle={International Conference on Learning Representations},
  volume={2026},
  pages={106613--106631},
  year={2026}
}

@article{xu2026mem,
  title={A-mem: Agentic memory for llm agents},
  author={Xu, Wujiang and Liang, Zujie and Mei, Kai and Gao, Hang and Tan, Juntao and Zhang, Yongfeng},
  journal={Advances in Neural Information Processing Systems},
  volume={38},
  pages={17577--17604},
  year={2026}
}

@article{chen2025telemem,
  title={Telemem: Building long-term and multimodal memory for agentic ai},
  author={Chen, Chunliang and Guan, Ming and Lin, Xiao and Li, Jiaxu and Lin, Luxi and Wang, Qiyi and Chen, Xiangyu and Luo, Jixiang and Sun, Changzhi and Zhang, Dell and others},
  journal={arXiv preprint arXiv:2601.06037},
  year={2025}
}

@article{wang2023voyager,
  title={Voyager: An open-ended embodied agent with large language models},
  author={Wang, Guanzhi and Xie, Yuqi and Jiang, Yunfan and Mandlekar, Ajay and Xiao, Chaowei and Zhu, Yuke and Fan, Linxi and Anandkumar, Anima},
  journal={arXiv preprint arXiv:2305.16291},
  year={2023}
}

@article{wang2024agent,
  title={Agent workflow memory},
  author={Wang, Zora Zhiruo and Mao, Jiayuan and Fried, Daniel and Neubig, Graham},
  journal={arXiv preprint arXiv:2409.07429},
  year={2024}
}

@article{lee2023explore,
  title={Explore, select, derive, and recall: Augmenting llm with human-like memory for mobile task automation},
  author={Lee, Sunjae and Choi, Junyoung and Lee, Jungjae and Wasi, Munim Hasan and Choi, Hojun and Ko, Steven Y and Oh, Sangeun and Shin, Insik},
  journal={arXiv preprint arXiv:2312.03003},
  year={2023}
}

@article{wang2025mobile,
  title={Mobile-agent-e: Self-evolving mobile assistant for complex tasks},
  author={Wang, Zhenhailong and Xu, Haiyang and Wang, Junyang and Zhang, Xi and Yan, Ming and Zhang, Ji and Huang, Fei and Ji, Heng},
  journal={arXiv preprint arXiv:2501.11733},
  year={2025}
}

@article{chen2026skilldroid,
  title={SkillDroid: Compile Once, Reuse Forever},
  author={Chen, Qijia and Bellucci, Andrea and Sun, Zhida and Jacucci, Giulio},
  journal={arXiv preprint arXiv:2604.14872},
  year={2026}
}

@article{huang2026memoharness,
  title={MemoHarness: Agent Harnesses That Learn from Experience},
  author={Huang, Yue and Wang, Wenjie and Bao, Han and Ma, Yuchen and Luo, Xiaonan and Nian, Yi and Zhuang, Haomin and Liu, Zheyuan and Zhao, Yue and Zhang, Xiangliang},
  journal={arXiv preprint arXiv:2607.14159},
  year={2026}
}

@article{pan2026natural,
  title={Natural-language agent harnesses},
  author={Pan, Linyue and Zou, Lexiao and Guo, Shuo and Ni, Jingchen and Zheng, Hai-Tao},
  journal={arXiv preprint arXiv:2603.25723},
  year={2026}
}

@article{lou2026autoharness,
  title={Autoharness: improving llm agents by automatically synthesizing a code harness},
  author={Lou, Xinghua and L{\'a}zaro-Gredilla, Miguel and Dedieu, Antoine and Wendelken, Carter and Lehrach, Wolfgang and Murphy, Kevin P},
  journal={arXiv preprint arXiv:2603.03329},
  year={2026}
}

@article{lee2026meta,
  title={Meta-harness: End-to-end optimization of model harnesses},
  author={Lee, Yoonho and Nair, Roshen and Zhang, Qizheng and Lee, Kangwook and Khattab, Omar and Finn, Chelsea},
  journal={arXiv preprint arXiv:2603.28052},
  year={2026}
}

@article{turing1950computing,
  title={Computing Machinery and Intelligence},
  author={Turing, Alan M.},
  journal={Mind},
  volume={LIX},
  number={236},
  pages={433--460},
  year={1950},
  doi={10.1093/mind/LIX.236.433}
}

@inproceedings{yang2023opro,
  title={Large language models as optimizers},
  author={Yang, Chengrun and Wang, Xuezhi and Lu, Yifeng and Liu, Hanxiao and Le, Quoc V and Zhou, Denny and Chen, Xinyun},
  booktitle={International Conference on Learning Representations},
  volume={2024},
  pages={12028--12068},
  year={2024}
}

@article{yuksekgonul2024textgrad,
  title={Textgrad: Automatic" differentiation" via text},
  author={Yuksekgonul, Mert and Bianchi, Federico and Boen, Joseph and Liu, Sheng and Huang, Zhi and Guestrin, Carlos and Zou, James},
  journal={arXiv preprint arXiv:2406.07496},
  year={2024}
}

@article{khattab2023dspy,
  title={Dspy: Compiling declarative language model calls into self-improving pipelines},
  author={Khattab, Omar and Singhvi, Arnav and Maheshwari, Paridhi and Zhang, Zhiyuan and Santhanam, Keshav and Vardhamanan, Sri and Haq, Saiful and Sharma, Ashutosh and Joshi, Thomas T and Moazam, Hanna and others},
  journal={arXiv preprint arXiv:2310.03714},
  year={2023}
}

@inproceedings{trirat2024automlagent,
  title={AutoML-Agent: A Multi-Agent LLM Framework for Full-Pipeline AutoML},
  author={Trirat, Patara and Jeong, Woomin and others},
  booktitle={Proceedings of the 42nd International Conference on Machine Learning},
  year={2025}
}

@article{sun2024card,
  title={A large language model-driven reward design framework via dynamic feedback for reinforcement learning},
  author={Sun, Shengjie and Liu, Runze and Lyu, Jiafei and Yang, Jing-Wen and Zhang, Liangpeng and Li, Xiu},
  journal={Knowledge-Based Systems},
  volume={326},
  pages={114065},
  year={2025},
  publisher={Elsevier}
}

@inproceedings{ma2023eureka,
  title={Eureka: Human-level reward design via coding large language models},
  author={Ma, Yecheng Jason and Liang, William and Wang, Guanzhi and Huang, De-An and Bastani, Osbert and Jayaraman, Dinesh and Zhu, Yuke and Fan, Jim and others},
  booktitle={International conference on learning Representations},
  volume={2024},
  pages={26516--26560},
  year={2024}
}

@article{zhang2025agenticrl,
  title={The landscape of agentic reinforcement learning for llms: A survey},
  author={Zhang, Guibin and Geng, Hejia and Yu, Xiaohang and Yin, Zhenfei and Zhang, Zaibin and Tan, Zelin and Zhou, Heng and Li, Zhongzhi and Xue, Xiangyuan and Li, Yijiang and others},
  journal={arXiv preprint arXiv:2509.02547},
  year={2025}
}

@inproceedings{wang2022selfinstruct,
  title={Self-instruct: Aligning language models with self-generated instructions},
  author={Wang, Yizhong and Kordi, Yeganeh and Mishra, Swaroop and Liu, Alisa and Smith, Noah A and Khashabi, Daniel and Hajishirzi, Hannaneh},
  booktitle={Proceedings of the 61st annual meeting of the association for computational linguistics (volume 1: long papers)},
  pages={13484--13508},
  year={2023}
}

@inproceedings{dong2024autoif,
  title={Self-play with execution feedback: Improving instruction-following capabilities of large language models},
  author={Dong, Guanting and Lu, Keming and Li, Chengpeng and Xia, Tingyu and Yu, Bowen and Zhou, Chang and Zhou, Jingren},
  booktitle={International Conference on Learning Representations},
  volume={2025},
  pages={39286--39313},
  year={2025}
}

@article{gunther2024cosmopedia,
  title={Cosmopedia: How to Create Large-Scale Synthetic Pretraining Data},
  author={G{\"{u}}nther, Moritz and others},
  journal={Hugging Face Blog},
  year={2024}
}

@inproceedings{lee2024llm2llm,
  title={Llm2llm: Boosting llms with novel iterative data enhancement},
  author={Lee, Nicholas and Wattanawong, Thanakul and Kim, Sehoon and Mangalam, Karttikeya and Shen, Sheng and Anumanchipalli, Gopala and Mahoney, Michael and Keutzer, Kurt and Gholami, Amir},
  booktitle={Findings of the Association for Computational Linguistics: ACL 2024},
  pages={6498--6526},
  year={2024}
}

@article{ye2026claw,
  title={Claw-Eval: Towards Trustworthy Evaluation of Autonomous Agents},
  author={Ye, Bowen and Li, Rang and Yang, Qibin and Liu, Yuanxin and Yao, Linli and Lv, Hanglong and Xie, Zhihui and An, Chenxin and Li, Lei and Kong, Lingpeng and others},
  journal={arXiv preprint arXiv:2604.06132},
  year={2026}
}

@article{lu2024aiscientist,
  title={Towards end-to-end automation of AI research},
  author={Lu, Chris and Lu, Cong and Lange, Robert Tjarko and Yamada, Yutaro and Hu, Shengran and Foerster, Jakob and Ha, David and Clune, Jeff},
  journal={Nature},
  volume={651},
  number={8107},
  pages={914--919},
  year={2026},
  publisher={Nature Publishing Group UK London}
}

@misc{anthropic2026modelreports,
  author={{Anthropic}},
  title={Anthropic's Transparency Hub},
  year={2026},
  howpublished={\url{https://www.anthropic.com/transparency}},
  note={Model reports and system cards for Claude Opus 5, Fable 5, and Opus 4.8. Accessed 2026-09-16}
}

@misc{google2026geminimodels,
  author={{Google}},
  title={{Gemini API}: Models},
  year={2026},
  howpublished={\url{https://ai.google.dev/gemini-api/docs/models}},
  note={Official model documentation, including Gemini 3.6 Flash and Gemini 3.1 Pro. Accessed 2026-09-16}
}

@misc{openai2026models,
  author={{OpenAI}},
  title={{OpenAI API}: Models},
  year={2026},
  howpublished={\url{https://developers.openai.com/api/docs/models}},
  note={Official model documentation, including GPT-6 Astra and GPT-5.6 Sol. Accessed 2026-09-16}
}

@misc{bytedance2026seed21,
  author={{ByteDance Seed Team}},
  title={{Seed2.1} Officially Released: Advancing {AI} Productivity},
  year={2026},
  month=jun,
  howpublished={\url{https://seed.bytedance.com/en/blog/seed2-1-officially-released-advancing-ai-productivity}}
}

@misc{bytedance2026seed20,
  author={{ByteDance Seed Team}},
  title={{Seed 2.0} Official Launch},
  year={2026},
  month=feb,
  howpublished={\url{https://seed.bytedance.com/blog/seed-2-0-official-launch}}
}

@misc{qwen37max,
  author={{Alibaba Cloud}},
  title={{Qwen3.7-Max}},
  year={2026},
  howpublished={\url{https://docs.modelstudio.console.alibabacloud.com/en/model-studio/qwen3-7-max}},
  note={Official model documentation. Accessed 2026-09-16}
}

@misc{qwen38,
  author={{Qwen Team}},
  title={{Qwen3.8-Max}: A New Bar for Coding and Cowork},
  year={2026},
  month=aug,
  howpublished={\url{https://qwen.ai/blog?id=qwen3.8}}
}

@article{glm5team2026glm5,
  author={{GLM-5 Team} and others},
  title={{GLM-5}: From Vibe Coding to Agentic Engineering},
  journal={arXiv preprint arXiv:2602.15763},
  year={2026},
  url={https://arxiv.org/abs/2602.15763}
}

@article{kimi2026k3,
  author={{Kimi Team} and others},
  title={{Kimi K3}: Open Frontier Intelligence},
  journal={arXiv preprint arXiv:2607.24653},
  year={2026},
  url={https://arxiv.org/abs/2607.24653}
}

@inproceedings{wu2026mcpmark,
  author={Wu, Zijian and Liu, Xiangyan and Zhang, Xinyuan and Chen, Lingjun and Meng, Fanqing and Du, Lingxiao and Zhao, Yiran and Zhang, Fanshi and Ye, Yaoqi and Wang, Jiawei and Wang, Zirui and Ni, Jinjie and Yang, Yufan and Xu, Arvin and Shieh, Michael Qizhe},
  title={{MCPMark}: A Benchmark for Stress-Testing Realistic and Comprehensive {MCP} Use},
  booktitle={International Conference on Learning Representations},
  year={2026},
  volume={2026},
  pages={79852--79895},
  url={https://proceedings.iclr.cc/paper_files/paper/2026/hash/8138d211ce8790fdfbeeeb9781838a37-Abstract-Conference.html}
}

@misc{mem02026research,
  author={{Mem0}},
  title={{Mem0} Research: {LoCoMo}, {LongMemEval} and {BEAM} Benchmarks},
  year={2026},
  howpublished={\url{https://mem0.ai/research}},
  note={Official evaluation protocols and reported results. Accessed 2026-09-16}
}

@inproceedings{maharana2024locomo,
  author={Maharana, Adyasha and Lee, Dong-Ho and Tulyakov, Sergey and Bansal, Mohit and Barbieri, Francesco and Fang, Yuwei},
  title={Evaluating Very Long-Term Conversational Memory of {LLM} Agents},
  booktitle={Proceedings of the 62nd Annual Meeting of the Association for Computational Linguistics (Volume 1: Long Papers)},
  year={2024},
  pages={13851--13870},
  doi={10.18653/v1/2024.acl-long.747},
  url={https://aclanthology.org/2024.acl-long.747/}
}

@article{wu2024longmemeval,
  author={Wu, Di and Wang, Hongwei and Yu, Wenhao and Zhang, Yuwei and Chang, Kai-Wei and Yu, Dong},
  title={{LongMemEval}: Benchmarking Chat Assistants on Long-Term Interactive Memory},
  journal={arXiv preprint arXiv:2410.10813},
  year={2024},
  url={https://arxiv.org/abs/2410.10813}
}

@article{tavakoli2025beam,
  author={Tavakoli, Mohammad and Salemi, Alireza and Ye, Carrie and Abdalla, Mohamed and Zamani, Hamed and Mitchell, J Ross},
  title={Beyond a Million Tokens: Benchmarking and Enhancing Long-Term Memory in {LLMs}},
  journal={arXiv preprint arXiv:2510.27246},
  year={2025},
  url={https://arxiv.org/abs/2510.27246}
}

@inproceedings{patil2025bfcl,
  author={Patil, Shishir G and Mao, Huanzhi and Yan, Fanjia and Ji, Charlie Cheng-Jie and Suresh, Vishnu and Stoica, Ion and Gonzalez, Joseph E.},
  title={The Berkeley Function Calling Leaderboard ({BFCL}): From Tool Use to Agentic Evaluation of Large Language Models},
  booktitle={Proceedings of the 42nd International Conference on Machine Learning},
  series={Proceedings of Machine Learning Research},
  volume={267},
  pages={48371--48392},
  publisher={PMLR},
  year={2025},
  url={https://proceedings.mlr.press/v267/patil25a.html}
}

@article{bandi2026mcpatlas,
  author={Bandi, Chaithanya and Dumitru, Razvan-Gabriel and Hertzberg, Ben and Agarwal, Divyansh and Boo, Geobio and Polakam, Tejas and Hassaan, Sami and Da, Jeff and Kim, HiJae and Gupta, Vipul and Sharma, Manasi and Park, Andrew and Dimakis, Martin and Montoya, Ernesto Gabriel Hernandez and Rambado, Dan and Salazar, Ivan and Cruz, Rafael and Rezaei, MohammadHossein and Rane, Chetan and Levin, Ben and Zhang, Daniel Yue and Kenstler, Brad and Liu, Bing},
  title={{MCP-Atlas}: A Large-Scale Benchmark for Tool-Use Competency with Real {MCP} Servers},
  journal={arXiv preprint arXiv:2602.00933},
  year={2026},
  url={https://arxiv.org/abs/2602.00933}
}

@misc{sierra2026tau3,
  author={{Sierra Research}},
  title={{\(\tau^3\)-bench}: Tool-Agent-User Interaction Benchmark},
  year={2026},
  howpublished={\url{https://github.com/sierra-research/tau2-bench}},
  note={Official repository and release documentation for the third benchmark version. Accessed 2026-09-16}
}

@article{li2025toolathlon,
  author={Li, Junlong and Zhao, Wenshuo and Zhao, Jian and Zeng, Weihao and Wu, Haoze and Wang, Xiaochen and Ge, Rui and Cao, Yuxuan and Huang, Yuzhen and Liu, Wei and Liu, Junteng and Su, Zhaochen and Guo, Yiyang and Zhou, Fan and Zhang, Lueyang and Michelini, Juan and Wang, Xingyao and Yue, Xiang and Zhou, Shuyan and Neubig, Graham and He, Junxian},
  title={The Tool Decathlon: Benchmarking Language Agents for Diverse, Realistic, and Long-Horizon Task Execution},
  journal={arXiv preprint arXiv:2510.25726},
  year={2025},
  url={https://arxiv.org/abs/2510.25726}
}

@misc{khandpur2025swemultilingual,
  author={Khandpur, Kabir and Lieret, Kilian and Jimenez, Carlos E. and Press, Ofir and Yang, John},
  title={{SWE-bench Multilingual}},
  year={2025},
  howpublished={\url{https://www.swebench.com/multilingual.html}},
  note={Official benchmark release}
}

@article{deng2025swepro,
  author={Deng, Xiang and Da, Jeff and Pan, Edwin and He, Yannis Yiming and Ide, Charles and Garg, Kanak and Lauffer, Niklas and Park, Andrew and Pasari, Nitin and Rane, Chetan and Sampath, Karmini and Krishnan, Maya and Kundurthy, Srivatsa and Hendryx, Sean and Wang, Zifan and Bharadwaj, Vijay and Holm, Jeff and Aluri, Raja and Zhang, Chen Bo Calvin and Jacobson, Noah and Liu, Bing and Kenstler, Brad},
  title={{SWE-Bench Pro}: Can {AI} Agents Solve Long-Horizon Software Engineering Tasks?},
  journal={arXiv preprint arXiv:2509.16941},
  year={2025},
  url={https://arxiv.org/abs/2509.16941}
}

@article{gema2024mmluredux,
  author={Gema, Aryo Pradipta and Leang, Joshua Ong Jun and Hong, Giwon and Devoto, Alessio and Mancino, Alberto Carlo Maria and Saxena, Rohit and He, Xuanli and Zhao, Yu and Du, Xiaotang and Madani, Mohammad Reza Ghasemi and Barale, Claire and McHardy, Robert and Harris, Joshua and Kaddour, Jean and van Krieken, Emile and Minervini, Pasquale},
  title={Are We Done with {MMLU}?},
  journal={arXiv preprint arXiv:2406.04127},
  year={2024},
  url={https://arxiv.org/abs/2406.04127}
}

@article{huang2023ceval,
  author={Huang, Yuzhen and Bai, Yuzhuo and Zhu, Zhihao and Zhang, Junlei and Zhang, Jinghan and Su, Tangjun and Liu, Junteng and Lv, Chuancheng and Zhang, Yikai and Lei, Jiayi and Fu, Yao and Sun, Maosong and He, Junxian},
  title={{C-Eval}: A Multi-Level Multi-Discipline Chinese Evaluation Suite for Foundation Models},
  journal={arXiv preprint arXiv:2305.08322},
  year={2023},
  url={https://arxiv.org/abs/2305.08322}
}

@article{zhou2023ifeval,
  author={Zhou, Jeffrey and Lu, Tianjian and Mishra, Swaroop and Brahma, Siddhartha and Basu, Sujoy and Luan, Yi and Zhou, Denny and Hou, Le},
  title={Instruction-Following Evaluation for Large Language Models},
  journal={arXiv preprint arXiv:2311.07911},
  year={2023},
  url={https://arxiv.org/abs/2311.07911}
}

@misc{anthropic2024mcp,
  author={{Anthropic}},
  title={Introducing the {Model Context Protocol}},
  year={2024},
  month=nov,
  howpublished={\url{https://www.anthropic.com/news/model-context-protocol}}
}

@article{wang2025roll,
  author={Wang, Weixun and Xiong, Shaopan and Chen, Gengru and Gao, Wei and Guo, Sheng and He, Yancheng and Huang, Ju and Liu, Jiaheng and Li, Zhendong and Li, Xiaoyang and others},
  title={Reinforcement Learning Optimization for Large-Scale Learning: An Efficient and User-Friendly Scaling Library},
  journal={arXiv preprint arXiv:2506.06122},
  year={2025},
  url={https://arxiv.org/abs/2506.06122}
}

@article{schulman2017ppo,
  author={Schulman, John and Wolski, Filip and Dhariwal, Prafulla and Radford, Alec and Klimov, Oleg},
  title={Proximal Policy Optimization Algorithms},
  journal={arXiv preprint arXiv:1707.06347},
  year={2017},
  url={https://arxiv.org/abs/1707.06347}
}

@article{shoeybi2019megatron,
  author={Shoeybi, Mohammad and Patwary, Mostofa and Puri, Raul and LeGresley, Patrick and Casper, Jared and Catanzaro, Bryan},
  title={{Megatron-LM}: Training Multi-Billion Parameter Language Models Using Model Parallelism},
  journal={arXiv preprint arXiv:1909.08053},
  year={2019},
  url={https://arxiv.org/abs/1909.08053}
}

@inproceedings{kwon2023vllm,
  author={Kwon, Woosuk and Li, Zhuohan and Zhuang, Siyuan and Sheng, Ying and Zheng, Lianmin and Yu, Cody Hao and Gonzalez, Joseph E. and Zhang, Hao and Stoica, Ion},
  title={Efficient Memory Management for Large Language Model Serving with {PagedAttention}},
  booktitle={Proceedings of the ACM SIGOPS 29th Symposium on Operating Systems Principles},
  year={2023},
  url={https://arxiv.org/abs/2309.06180}
}

@inproceedings{moritz2018ray,
  author={Moritz, Philipp and Nishihara, Robert and Wang, Stephanie and Tumanov, Alexey and Liaw, Richard and Liang, Eric and Elibol, Melih and Yang, Zongheng and Paul, William and Jordan, Michael I. and Stoica, Ion},
  title={{Ray}: A Distributed Framework for Emerging {AI} Applications},
  booktitle={13th USENIX Symposium on Operating Systems Design and Implementation (OSDI 18)},
  year={2018},
  pages={561--577},
  publisher={USENIX Association},
  url={https://www.usenix.org/conference/osdi18/presentation/moritz}
}

@incollection{good1965speculations,
  title={Speculations Concerning the First Ultraintelligent Machine},
  author={Good, I. J.},
  booktitle={Advances in Computers},
  editor={Alt, Franz L. and Rubinoff, Morris},
  volume={6},
  pages={31--88},
  publisher={Academic Press},
  year={1965}
}
}

\newpage
{\Large\textbf{Appendix}}
\appendix

\section{Harness Configuration and Memory Details}
\label{app:harness-details}

This appendix expands the deployment mechanisms summarized in Section~\ref{sec:harness-engineering}. The Harness is the common runtime that assembles context, connects the Planner Model to external resources, relays structured execution feedback, and retains traces for diagnosis. The Scenario Adapter and Persistent Memory Manager are internal components of this runtime rather than independent runtime systems.

\subsection{Scenario-Adaptation Configuration}
\label{app:scenario-adapter}

The Scenario Adapter converts deployment-time tool availability into compact operational guidance. Offline, it compiles tool schemas, capability-domain Skill definitions, a complete tool-to-Skill mapping, and guidance constraints into a versioned configuration. Each build records a digest of the target tool schemas, checks coverage of the deployed tool set, and resolves ambiguous mappings through explicit overrides. The resulting artifact contains a tool-to-Skill index, reusable domain guidance, global execution rules, candidate-specific routing notes, multi-turn instructions, priority rules, and configurable limits on injected content.

At request time, the Adapter receives the tools already exposed to the planner; it does not discover additional resources. It maps these candidate tools to the corresponding Skills and filters each guidance block against the request-visible tool set. General rules are retained, instructions tied exclusively to unavailable tools are removed, and tool-family lists are rewritten to include only reachable resources. Candidate-specific routing notes and conditional multi-turn guidance are then added before the result is incorporated into the active context. Requests that explicitly expose Skill tools receive a generic Skill-use protocol instead of candidate-derived guidance, preventing the Adapter from revealing or bypassing the intended Skill-selection process.

The planner remains responsible for interpreting the request, resolving conflicts, selecting actions, and replanning after feedback. The Scenario Adapter neither executes actions nor persists device state. Agent-backed capabilities and atomic tools enter the same structured action interface, while Skills provide procedural guidance for composing their use rather than acting as opaque commands.

\subsection{Persistent Memory Lifecycle}
\label{app:persistent-memory}

The Persistent Memory Manager stores cross-interaction information outside the model in an inspectable workspace. It distinguishes a compact user model for stable preferences and behavioral requirements, episodic records for detailed observations and dialogue evidence, curated long-term memory for consolidated facts and lessons, prospective memory for future intentions with explicit activation conditions, and review records that document maintenance without becoming authoritative memories. These layers differ in both function and access: compact validated information may be made readily available, whereas detailed or less trusted evidence requires retrieval under the current task.

Memory formation begins with observations and dialogue evidence captured during interaction. Before a long context is compressed, a persistence step records important unstored requirements, decisions, and state changes. Each memory unit links a structured summary to source metadata and a lossless dialogue record. The summary supports efficient retrieval; metadata records provenance, speaker, turn, and order; and the original record permits direct verification of numbers, dates, quotations, contradictions, and changing facts. Invalid summaries are treated as ingestion failures rather than silently replaced, keeping ingestion success, summary quality, and source fidelity separately observable.

Information remains episodic until a staged process establishes that it is useful and sufficiently trustworthy for longer-term use. Deterministic checks first enforce eligibility, provenance, and trust requirements; bounded model-assisted processing is then used to organize related observations, remove duplicates, identify recurring evidence, and express accepted information concisely. Consolidation merges duplicates with existing memory, preserves source links, and marks obsolete entries as superseded when newer evidence changes the current state. Recalled material is tagged so that it is not re-ingested as new evidence, preventing repeated retrieval from artificially increasing its apparent importance.

Retrieval combines semantic and keyword matching with relevance, recency, importance, and diversity controls. Query structure determines how evidence is assembled: temporal questions require event ordering, cross-session aggregation requires entity and action deduplication, and questions about updated facts require explicit resolution of superseding evidence. For long histories, retrieval may proceed through high-recall search, construction of a query-specific evidence ledger, source verification, and focused re-examination. Retrieved Memory remains evidence supplied to the planner; it does not authorize or execute an action.

Provenance and lifecycle controls bound the authority of stored content. User-provided, agent-derived, externally sourced, and system-generated information remain distinguishable, and content cannot promote its own trust level. Prospective items carry activation conditions, scope, expiry, and completion status. Consolidation reports record additions, merges, and superseded entries, while earlier versions are retained before substantial revisions. If semantic search or model-assisted consolidation is unavailable, the system can fall back to simpler retrieval, retain information in episodic form, or postpone maintenance without blocking the primary task.

\subsection{Memory Evaluation and Reproducibility}
\label{app:memory-evaluation}

LoCoMo, LongMemEval, and BEAM are evaluated through a common hierarchy of samples, sessions, dialogue turns, and questions while retaining their original task semantics and scoring rules. We compare models paired with the Persistent Memory Manager against their corresponding model-only settings. Memory ingestion, answer generation, aggregation, and error handling follow the same pipeline, and answer generation, retrieval, and automated judging are assigned separate roles to reduce coupled model bias.

Retrieval integrity is part of a valid execution. If the planner invokes a retrieval operation, the trace must contain at least one successful result. A failed attempt followed by a successful one is recorded as recovery; traces in which every attempt fails remain eligible for retry. Timeouts, context overflows, embedding failures, and retrieval-service errors are recorded as infrastructure failures rather than model errors. The evaluation pipeline records protocol accuracy over valid executions, end-to-end accuracy over all samples, and valid-execution coverage separately so that model quality is not conflated with runtime reliability; the aggregate scores in Table~\ref{tab:memory-benchmarks} follow each benchmark's stated scoring protocol.

Reproducibility is supported through sample-level isolation, configuration fingerprints, ingestion snapshots, and per-question checkpoints. Each evaluation unit has an independent session, index, and result space, enabling interrupted runs to resume from a known state. The retained artifact chain includes dataset manifests, memory records, retrieval traces, model calls, judge outputs, configurations, and aggregate metrics. These records make it possible to distinguish failures caused by memory construction, missing retrieval evidence, answer generation, automated scoring, or infrastructure.

\section{Infrastructure Details}
\label{app:training-details}
Our training infrastructure supports the main stages of mobile-agent development, including supervised fine-tuning, long-context training, reinforcement learning, and on-policy distillation. It combines distributed model training and rollout generation with a shared environment management layer. The same system can support both dense and Mixture-of-Experts (MoE) models, as well as tasks running in sandboxes, simulators, and real mobile devices.

\paragraph{Agentic RL.}
We build our agentic reinforcement learning system on Roll~\citep{wang2025roll}. Megatron-Core~\citep{shoeybi2019megatron} is used for distributed policy training, while vLLM~\citep{kwon2023vllm} provides efficient rollout inference. Ray~\citep{moritz2018ray} coordinates training, inference, environment interaction, and reward computation across the cluster. This design allows each part of the pipeline to use resources according to its own workload.

Training and rollout generation run asynchronously on separate groups of GPUs. Rollout workers continue to interact with environments and generate trajectories while learner workers update the policy. This separation reduces idle time and improves system efficiency compared with a fully synchronous pipeline. It also makes the allocation of training and inference resources easier to adjust for different models and tasks.

The system supports long-context reinforcement learning for large MoE models through several forms of distributed parallelism. Training configurations are selected based on model size, context length, memory use, and communication cost. For example, context parallelism distributes long sequences across devices, while expert parallelism spreads MoE experts across the training cluster. Group-based policy optimization is used to compare multiple trajectories for each task and provide a more stable learning signal.

The system also checks that parallel groups, rollout batches, and model partitions are compatible before training begins. These checks help distribute samples and tokens evenly across devices and prevent invalid distributed configurations.

\paragraph{Environment Management.}
\label{app:environment-management}
Section~\ref{sec:environment-management} introduces the division of responsibilities between Roll and ROCK. At the implementation level, the same environment layer is reused for evaluation, supervised data collection, and reinforcement learning, and it supports general-agent and code-agent tasks through backend-specific adapters. This reuse allows sandboxed tasks, simulated mobile tasks, and real-device interactions to enter a shared training pipeline without requiring identical execution engines.

During online training, Roll manages distributed rollout generation, model updates, and training-resource scheduling, while ROCK manages environment instances, device sessions, and their execution lifecycles. Environment validation and trajectory filtering are applied before samples are admitted to training. These checks preserve the separation between distributed model computation and backend-specific execution while keeping the resulting trajectories compatible with the common data and optimization pipeline.

\clearpage
\section{Additional Mobile-Planning Case Studies}
\label{app:mobile-planning-cases}

These six trajectories extend Section~\ref{sec:behavioral-analysis} in a progression from procedural dependencies, through scoped revisions, to context use across user turns. They show Qwen-Planner-Agent recovering complete procedures, sequencing actions with interacting effects, limiting device and memory revisions to their intended scope, and reusing information across dialogue turns. Each task is completed with supporting tool returns or stored state. Each figure retains all tool-call rounds of the current interaction. Earlier turns in the two dialogue examples provide the interaction history, and repeated letter and H indexes connect prior requirements or observations to the highlighted decisions.

\subsection{Procedural Dependencies}

\begin{figure}[!htbp]
    \centering
    \captionsetup{font=small,labelfont=bf,skip=5pt}
    \includegraphics[width=\linewidth,height=0.84\textheight,keepaspectratio]{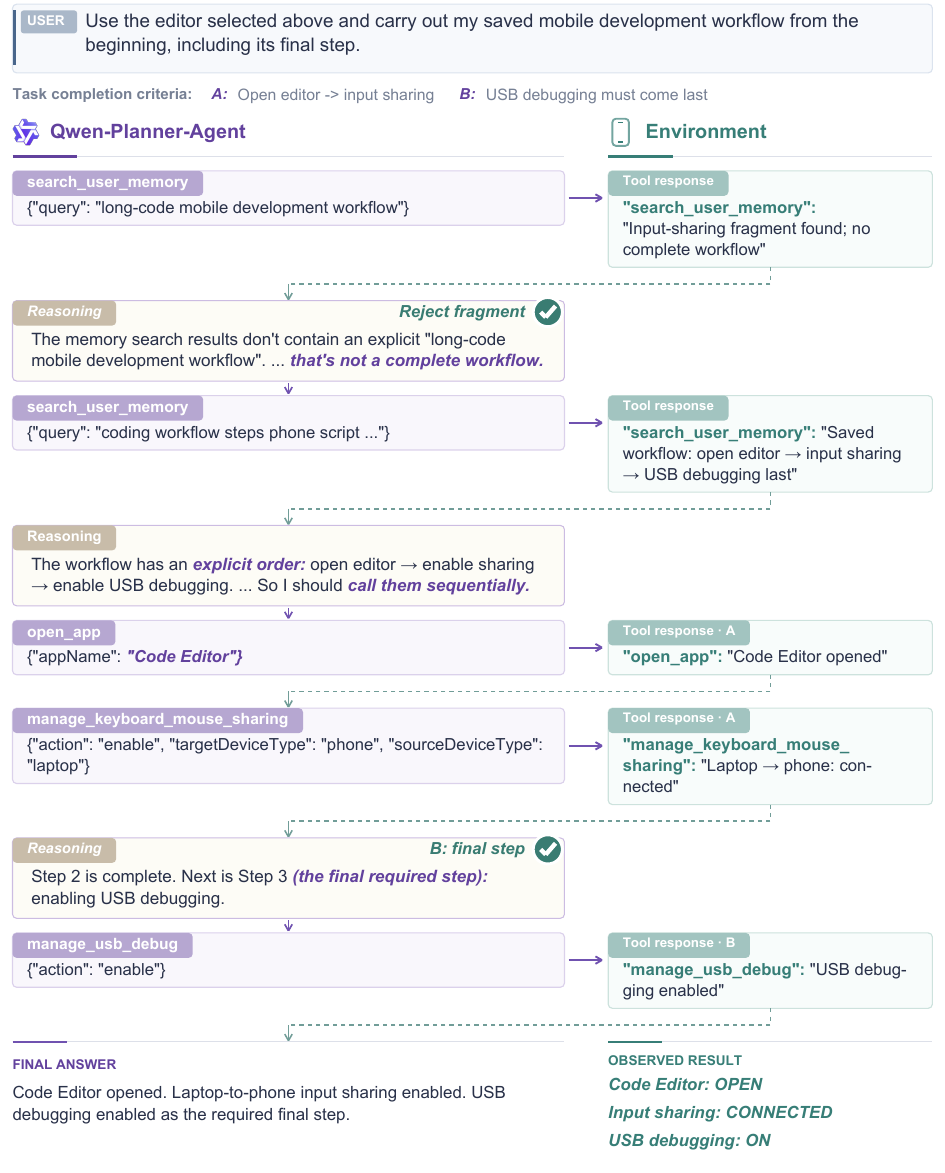}
    \caption{\textbf{Memory-guided recovery of an ordered procedure.} Qwen-Planner-Agent recognizes that the first memory is incomplete and retrieves the missing workflow. It then preserves the recovered dependency order: editor launch and input sharing (A) precede USB debugging (B), with successful returns confirming all three actions.}
    \label{fig:vis_long_coding_workflow}
\end{figure}

\Needspace{6\baselineskip}
\paragraph{Recovering complete procedures through targeted retrieval.} Figure~\ref{fig:vis_long_coding_workflow} examines whether the planner can detect that a retrieved memory is insufficient for the requested procedure. The first result covers input sharing but omits the complete coding setup. Qwen-Planner-Agent identifies the missing information, retrieves the full workflow, and preserves its order through editor launch, input sharing, and USB debugging. The A/B markers separate the setup prerequisites from the final debugging action, and each returned state is incorporated before the next dependent call.

\begin{figure}[!htbp]
    \centering
    \captionsetup{font=small,labelfont=bf,skip=5pt}
    \includegraphics[width=\linewidth,height=0.84\textheight,keepaspectratio]{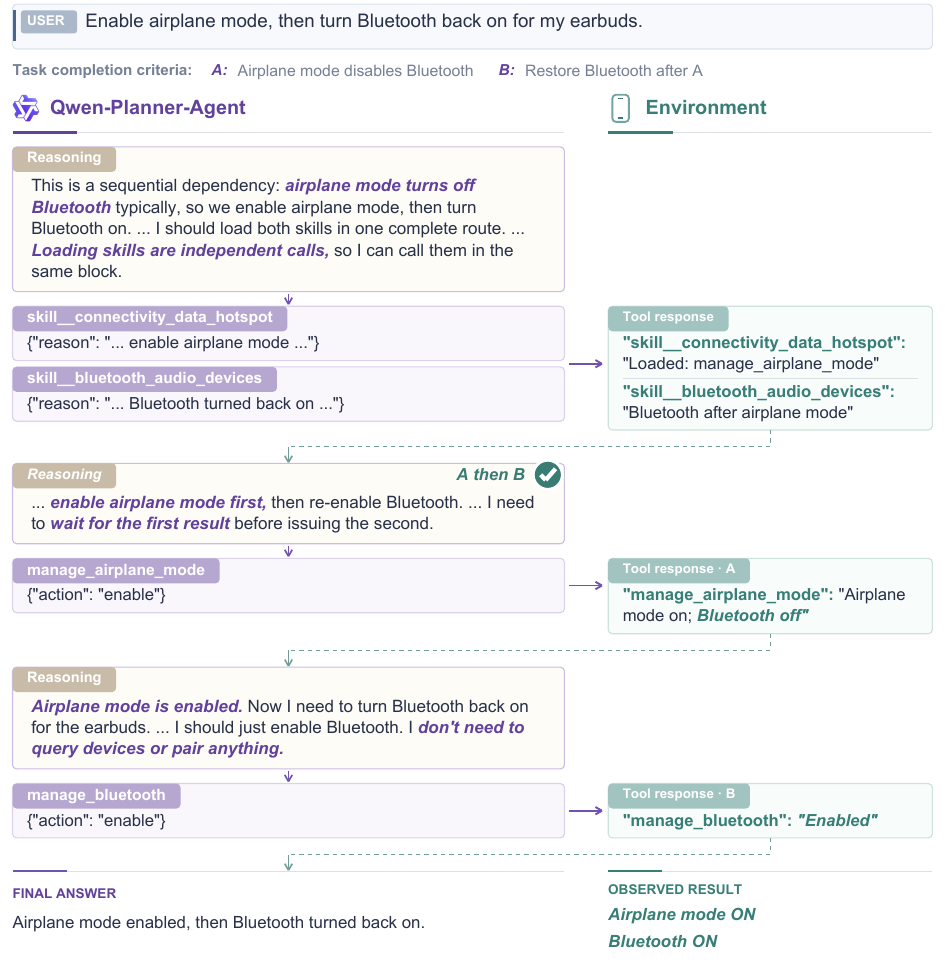}
    \caption{\textbf{Skill-guided planning over interacting state changes.} Qwen-Planner-Agent loads the relevant Skills together, then enables airplane mode (A) before restoring Bluetooth (B). This order accounts for the first action's side effect and leaves both requested settings enabled.}
    \label{fig:vis_mobile_skill_1}
\end{figure}

\Needspace{6\baselineskip}
\paragraph{Composing Skills under execution dependencies.} Figure~\ref{fig:vis_mobile_skill_1} separates parallelizable preparation from state-dependent execution. The connectivity and Bluetooth Skills can be loaded together, but their device actions cannot be applied in arbitrary order because enabling airplane mode disables Bluetooth. Qwen-Planner-Agent therefore waits for the airplane-mode transition (A) before restoring Bluetooth (B). The final returns verify both requested settings, showing that the planner accounts for cross-tool side effects rather than treating the calls as independent.

\subsection{Scoped Device and Memory Revisions}

\begin{figure}[!htbp]
    \centering
    \captionsetup{font=small,labelfont=bf,skip=5pt}
    \includegraphics[width=\linewidth,height=0.84\textheight,keepaspectratio]{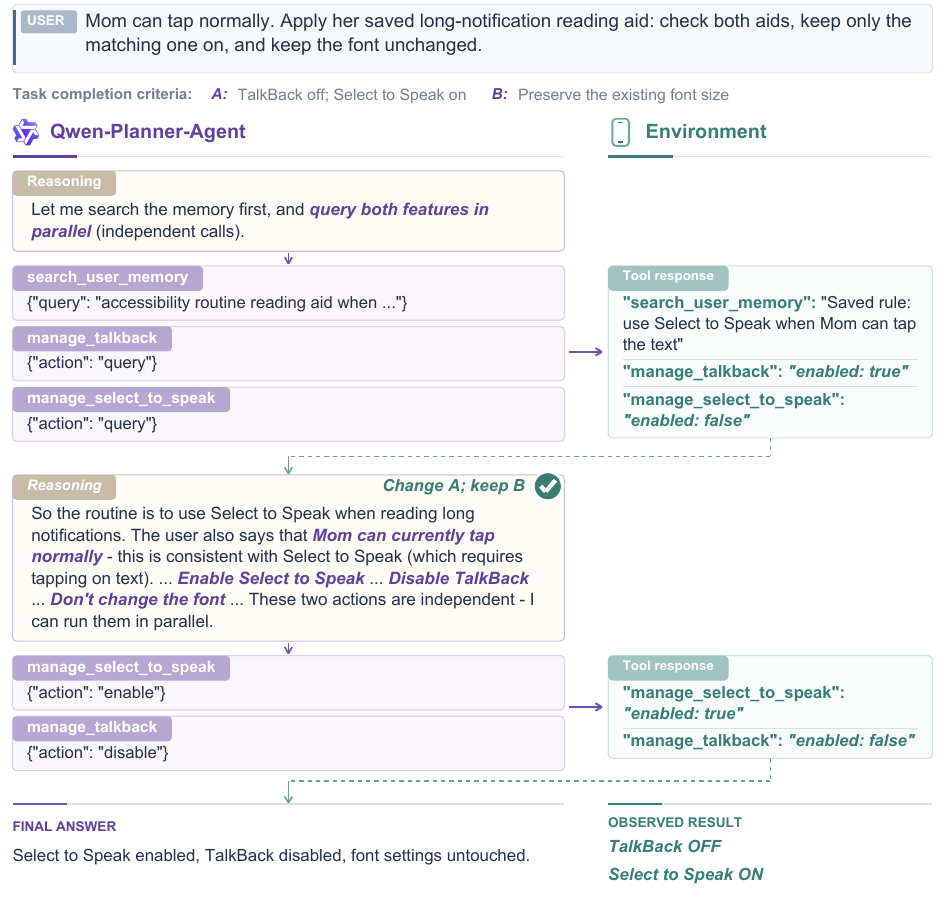}
    \caption{\textbf{Memory-grounded correction within an explicit scope.} Qwen-Planner-Agent observes both reading-aid states, enables Select to Speak, and disables TalkBack (A). It leaves the font setting unchanged (B), realizing the stored preference with only the required state changes.}
    \label{fig:vis_mobile_state_correction}
\end{figure}

\Needspace{6\baselineskip}
\paragraph{Correcting device state within explicit constraints.} Figure~\ref{fig:vis_mobile_state_correction} tests whether a stored preference can be translated into the smallest valid state change. The observed configuration has TalkBack enabled and Select to Speak disabled, while the preference requires the reverse and explicitly preserves the font setting. Qwen-Planner-Agent queries the relevant states, applies the two necessary accessibility changes (A), and leaves the font configuration untouched (B). This observation--delta--verification pattern grounds memory in device state while respecting an explicit non-interference constraint.

\begin{figure}[!htbp]
    \centering
    \captionsetup{font=small,labelfont=bf,skip=5pt}
    \includegraphics[width=\linewidth,height=0.84\textheight,keepaspectratio]{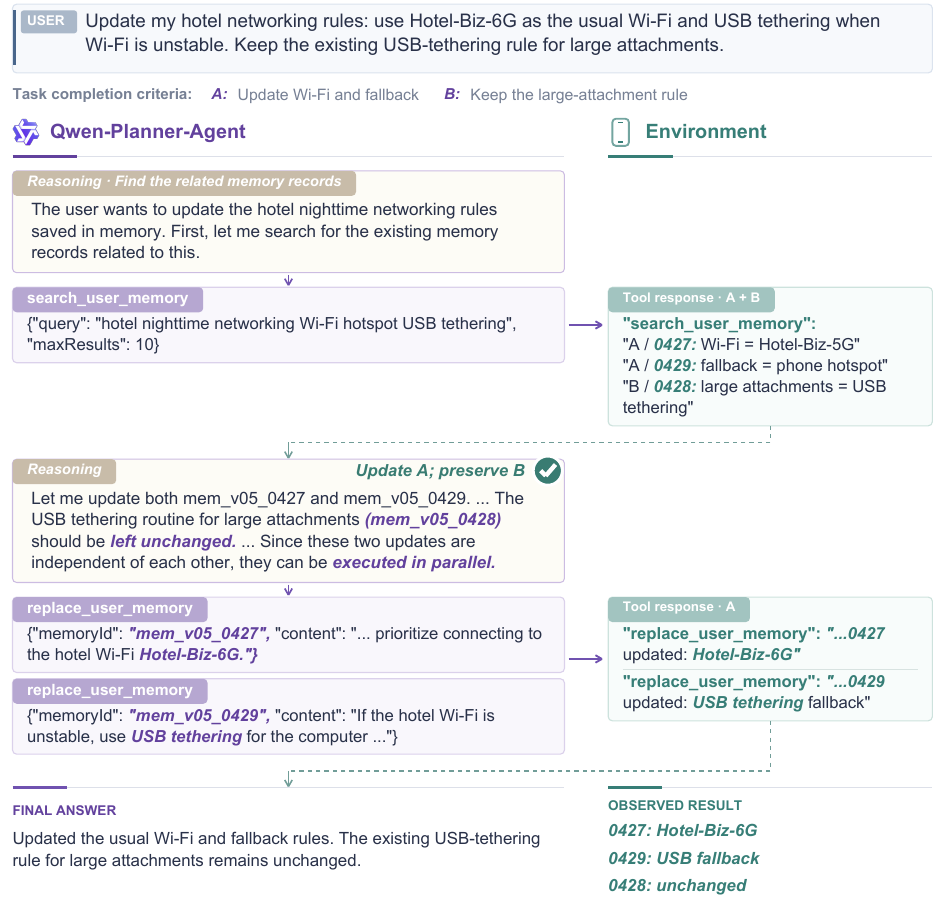}
    \caption{\textbf{Coordinated revision of persistent memory.} Qwen-Planner-Agent updates records 0427 and 0429 for the new Wi-Fi and USB fallback (A), while retaining the large-attachment rule in 0428 (B). The same record identifiers connect retrieval to the two targeted writes; the final memory state confirms that the third record is unchanged.}
    \label{fig:memory_coordinated_revision}
\end{figure}

\Needspace{6\baselineskip}
\paragraph{Updating persistent memory within the requested scope.} Related memories may require different treatment under the same request. In Figure~\ref{fig:memory_coordinated_revision}, Qwen-Planner-Agent identifies three hotel-networking records, updates the usual Wi-Fi and the fallback for unstable connections, and retains the existing large-attachment rule. It issues the two independent replacements together using their original record identifiers. The final memory state confirms both revisions and the preserved third record, showing how a compound preference change can be applied without duplicating memories or broadening the edit.

\clearpage
\subsection{Constraint Use Across Dialogue Turns}

\begin{figure}[!htbp]
    \centering
    \captionsetup{font=small,labelfont=bf,skip=5pt}
    \includegraphics[width=\linewidth,height=0.84\textheight,keepaspectratio]{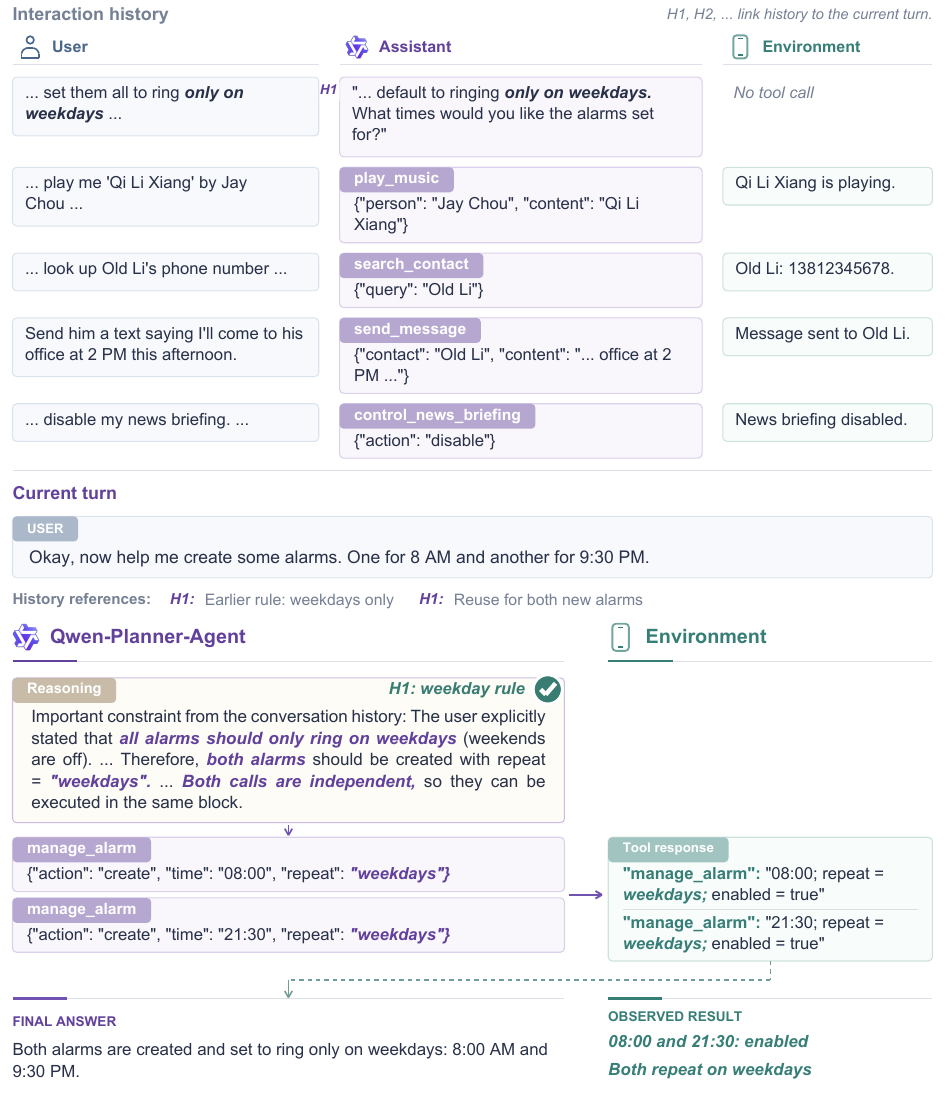}
    \caption{\textbf{Constraint-preserving tool use across dialogue turns.} H1 links the earlier weekday-only requirement to two alarm requests after four intervening topics. Qwen-Planner-Agent carries the retained constraint into both tool calls, and both returned records confirm weekday repetition.}
    \label{fig:vis_multiturn_weekday_constraint}
\end{figure}

\Needspace{6\baselineskip}
\paragraph{Retaining task constraints across topic changes.} Figure~\ref{fig:vis_multiturn_weekday_constraint} places four intervening turns between a shared weekday-only alarm requirement and a later request specifying just two times. Qwen-Planner-Agent applies the earlier constraint (H1) to both creation calls. The returned records confirm weekday repetition, showing that the constraint remains effective across topic changes without being restated.

\begin{figure}[!htbp]
    \centering
    \captionsetup{font=small,labelfont=bf,skip=5pt}
    \includegraphics[width=\linewidth,height=0.84\textheight,keepaspectratio]{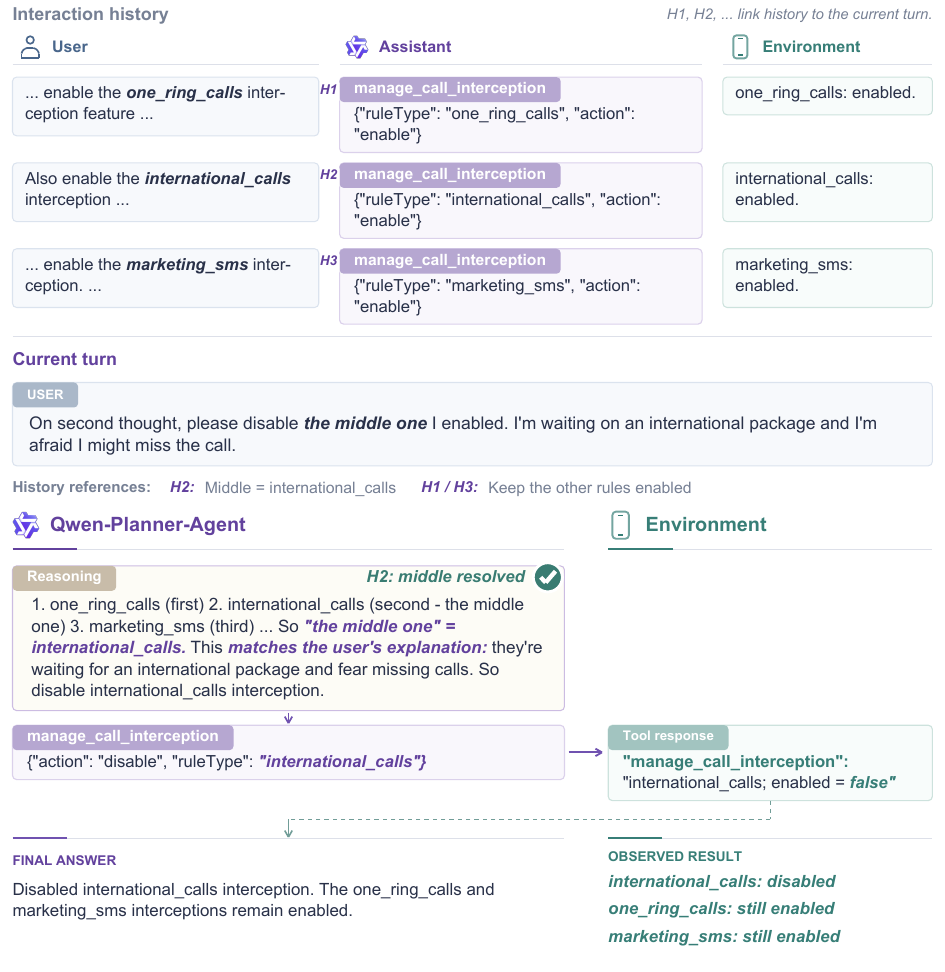}
    \caption{\textbf{Reference-grounded tool use over accumulated device state.} Qwen-Planner-Agent resolves ``the middle one'' as international-call interception (H2), consistent with the user's explanation. It disables only that rule while preserving one-ring-call and marketing-message interception (H1 and H3).}
    \label{fig:vis_multiturn_scoped_revision}
\end{figure}

\Needspace{6\baselineskip}
\paragraph{Resolving dialogue references for targeted state revisions.} In Figure~\ref{fig:vis_multiturn_scoped_revision}, the user asks to disable ``the middle one'' after establishing three interception rules. Qwen-Planner-Agent combines interaction order with the international-delivery explanation to identify international-call interception (H2). The final state confirms that only this rule is disabled, preserving one-ring-call and marketing-message interception (H1 and H3).

\end{document}